\documentclass{article}

\usepackage{PRIMEarxiv}

\usepackage[utf8]{inputenc} 
\usepackage[T1]{fontenc}    
\usepackage{hyperref}       
\usepackage{url}            
\usepackage{booktabs}       
\usepackage{amsfonts}       
\usepackage{nicefrac}       
\usepackage{microtype}      
\usepackage{lipsum}
\usepackage{fancyhdr}       
\usepackage{graphicx}       
\graphicspath{{media/}}     
\usepackage{multirow}
\usepackage{multicol}
\usepackage{lscape}
\usepackage{tablefootnote}
\usepackage{subfigure}
\usepackage{float}
\usepackage{array}
\title{UAV (Unmanned Aerial Vehicles): Diverse Applications of UAV Datasets in Segmentation, Classification, Detection, and Tracking
}

\author{
 Md. Mahfuzur Rahman \\
 Software Engineer \\
 Silicon Orchard Research and Analytics Lab \\
 Dhaka, Bangladesh\\
 \texttt{mahim1066@gmail.com} \\
 \And
 Sunzida Siddique \\
 Dept. of Computer Science \\
 Daffodil International University \\
 Dhaka, Bangladesh\\
 \texttt{sunzida15-9667@diu.edu.bd} \\
 \And
 Marufa Kamal \\
 Dept. of CSE \\
 BRAC University \\
 Dhaka, Bangladesh\\
 \texttt{marufa.kamal1@g.bracu.ac.bd} \\
 \And
 Rakib Hossain Rifat \\
 Dept. of Computer Science \\
 Texas Tech University \\
 Lubbock, TX 79409\\
 \texttt{rrifat@ttu.edu} \\
 \And
 Kishor Datta Gupta \\
 Cyber Physical Systems \\
 Clark Atlanta University \\
 Atlanta, GA\\
 \texttt{kgupta@cau.edu} \\
}

\newcolumntype{C}[1]{>{\centering\arraybackslash}p{#1}}

\begin{document}
\maketitle

\begin{abstract}
Unmanned Aerial Vehicles (UAVs), have greatly revolutionized the process of gathering and analyzing data in diverse research domains, providing unmatched adaptability and effectiveness. This paper presents a thorough examination of Unmanned Aerial Vehicle (UAV) datasets, emphasizing their wide range of applications and progress. UAV datasets consist of various types of data, such as satellite imagery, images captured by drones, and videos. These datasets can be categorized as either unimodal or multimodal, offering a wide range of detailed and comprehensive information. These datasets play a crucial role in disaster damage assessment, aerial surveillance, object recognition, and tracking. They facilitate the development of sophisticated models for tasks like semantic segmentation, pose estimation, vehicle re-identification, and gesture recognition. By leveraging UAV datasets, researchers can significantly enhance the capabilities of computer vision models, thereby advancing technology and improving our understanding of complex, dynamic environments from an aerial perspective. This review aims to encapsulate the multifaceted utility of UAV datasets, emphasizing their pivotal role in driving innovation and practical applications in multiple domains.
\end{abstract}


\section{Introduction}
Unmanned Aerial Vehicles (UAVs)\cite{b16}, commonly referred to as drones, have revolutionized the way we collect and analyze data from above, offering unparalleled versatility and efficiency across various research fields. This review paper aims to explore the "Multiple Uses of UAV Datasets" by examining the diverse applications and advancements facilitated by these datasets. UAV datasets encompass a wide array of data types, including satellite imagery, drone-captured images, and videos, as well as images from other aerial vehicles like helicopters. These datasets can be unimodal, focusing on a single type of data, or multimodal, integrating multiple data types to provide deeper, more comprehensive insights.

UAV datasets have proven to help assess disaster damage because they allow for the classification of damage from natural disasters using sophisticated semantic segmentation and annotation techniques. By training computer vision models with these datasets, researchers can automate the aerial scene classification of disaster events, significantly enhancing response and recovery efforts. The ability to extract information and detect objects from UAV-captured data is pivotal for tasks such as action recognition, where human behavior is analyzed from aerial imagery, including recognizing aerial gestures and classifying disaster events.

A critical application of UAV datasets lies in 'Aerial Surveillance'\cite{b18}, which supports advanced research at the intersection of computer vision, robotics, and surveillance. These datasets are used for event recognition in aerial videos, aiding in the monitoring of urban environments and traffic systems. The use of pre-trained models and transfer learning techniques further amplifies the utility of UAV datasets, allowing for the rapid deployment of sophisticated models for event recognition and tracking.

In the context of urban surveillance, UAV datasets enhance object recognition capabilities by providing comprehensive views from both top-down and side perspectives. This facilitates tasks such as categorization, verification, object detection, and tracking of individuals and vehicles. Moreover, UAV datasets contribute significantly to understanding and managing forest ecosystems by addressing the challenge of segmenting individual trees, which is crucial for sustainable forest management.

The versatility of UAV datasets extends to various domains, including developing speech recognition systems for UAV control using video capture and object tracking in low-light conditions, which is essential for night-time surveillance operations. Innovative UAV designs, such as bionic drones with flapping wings, have also led to specialized video datasets used for single object tracking (SOT)\cite{b21}\cite{b22}, demonstrating the broad scope and potential of UAV datasets in enhancing real-time object tracking under varying lighting conditions.

Overall, UAV datasets represent a cornerstone for cutting-edge research and practical applications across multiple disciplines. This review will delve into the specific uses and benefits of these datasets, highlighting their role in advancing technology and improving our understanding of complex, dynamic environments from an aerial perspective.

The subsequent sections provide a comprehensive exposition of the contributions made by our study, which can be stated as follows:
\begin{itemize}
    \item Our study is driven by the increasing importance of UAV datasets in several research domains such as object detection, traffic monitoring, action identification, surveillance in low-light conditions, single object tracking, and forest segmentation utilizing point cloud or LiDAR point process modeling. Through an in-depth analysis of current datasets, their uses, and prospects, this paper intends to provide valuable insights that will assist researchers in harnessing these resources for creative solutions. Furthermore, they will acquire knowledge of existing constraints and prospective opportunities, enhancing their research endeavors.
    \item We conducted an extensive analysis of a dataset consisting of 15 Unmanned Aerial Vehicles (UAVs), showcasing its diverse applications in research.
    \item  We emphasized the applications and advancements of several novel methods utilizing these datasets based on unmanned aerial vehicles (UAVs).
    \item Our study also delved into the potential for future research and the feasibility of utilizing these UAV datasets, engaging in in-depth discussions on these topics.
\end{itemize}

\section{Literature Review}
An unmanned aircraft or UAV, functions without a human pilot on board and can be operated remotely by a human controller or independently by onboard computers. Drones are a common term used to describe UAVs. Drones are employed for various purposes, including surveillance, aerial photography, agriculture, environmental monitoring, and military operations. However, within the UAV dataset context, the term encompasses more than just drones. UAV datasets encompass not only drone image and video datasets, but also include satellite imagery. Table\ref{tab:dataset_summary_1} and \ref{tab:dataset_summary_2} shows the summary of the literature review performed. 

These papers were reviewed to determine the definition and range of applications of UAVs in computer vision.

\subsection{RescueNet}
Maryam Rahnemoonfar, Tashnim Chowdhury, and Robin Murphy presented the RescueNet\cite{b1} dataset in their paper, which focuses on post-disaster scene understanding using UAV imagery. The dataset contains high-resolution images with detailed pixel-level annotations for ten classes of objects, including buildings, roads, pools, and trees, which were collected by sUAVs following Hurricane Michael. The authors employed state-of-the-art segmentation models like Attention UNet\cite{b19}, PSPNet\cite{b20}, and DeepLabv3\cite{b23}, achieving superior performance with attention-based and transformer-based methods. The findings demonstrated RescueNet's effectiveness in improving damage assessment and response strategies, with transfer learning outperforming other datasets like FloodNet\cite{b17}. The dataset was observed to have limited generalization to other domains and to require a time-consuming annotation process, despite its detailed annotations.

\subsection{UAV-Human}
Tianjiao Li et al. developed the UAV-Human\cite{b2} dataset, a comprehensive benchmark for improving human behavior understanding with UAVs. The dataset contains 67,428 multi-modal video sequences with 119 subjects for action recognition, 22,476 frames for pose estimation, 41,290 frames for person re-identification with 1,144 identities, and 22,263 frames for attribute recognition, all captured over three months in various urban and rural locations under varying conditions. The data encompasses RGB videos, depth maps, infrared sequences, and skeleton data. The authors used methods such as HigherHRNet\cite{b24}, AlphaPose\cite{b25}, and the Guided Transformer I3D framework to recognize actions while addressing fisheye video distortions\cite{b26}\cite{b27} and leveraging multiple data modalities. The results demonstrated the dataset's effectiveness in improving action recognition, pose estimation, and re-identification tasks, with models showing significant performance improvements. The UAV-Human dataset stands out as a reliable benchmark, encouraging the creation of more effective UAV-based human behavior analysis algorithms. 

\subsection{AIDER}
Christos Kyrkou and Theocharis Theocharides introduced the AIDER\cite{b6} dataset, which is intended for disaster event classification using UAV aerial images. The dataset contains 2,565 images of Fire/Smoke, Flood, Collapsed Building/Rubble, Traffic Accidents, and Normal cases, which were manually collected from various sources, mainly from UAVs. To increase variability and combat overfitting, images were randomly augmented with rotations, translations, and color shifting. The paper presents ERNet, a lightweight CNN designed for efficient classification on embedded UAV platforms. ERNet, which uses components from architectures such as VGG16\cite{b28}, ResNet\cite{b29}, and MobileNet\cite{b30}, incorporates early downsampling to reduce computational costs. When tested on both embedded platforms attached to UAVs and desktop CPUs, ERNet achieved almost perfect accuracy (90\%) while running three times faster on embedded platforms. This showed that it is a good choice for real-time applications that do not need a lot of memory. The study emphasizes the benefits of combining ERNet with other detection algorithms to improve situational awareness in emergency response. 

\subsection{AU-AIR}
In their paper Ilker Bozcan and Erdal Kayacan present the AU-AIR\cite{b3} dataset, a comprehensive UAV dataset designed for traffic surveillance. The dataset comprises 32,823 labeled video frames with annotations for eight traffic-related object categories, along with multi-modal data including GPS coordinates, altitude, IMU data\cite{b31}, and velocity. To establish a baseline for real-time performance in UAV applications, the authors train and evaluate two mobile object detectors on this dataset: YOLOv3-Tiny\cite{b32} and MobileNetv2-SSDLite\cite{b33}. The findings highlight the difficulties of object detection in aerial images, emphasizing the importance of datasets tailored to mobile detectors. The study highlights the dataset's potential for furthering research in computer vision, robotics, and aerial surveillance, while also acknowledging limitations and suggesting future improvements for broader applicability.

\subsection{ERA}
Lichao Mou et al. introduced the ERA\cite{b7} dataset, a comprehensive collection of 2,864 labeled video snippets for 24 event classes and 1 normal class, designed for event recognition in UAV videos. The videos, sourced from YouTube, are 5 seconds long, 640×640 pixels, and run at 24 fps, ensuring a diverse dataset that includes both high-quality and extreme condition footage. The paper employs various deep learning models, including VGG-16, ResNet-50, DenseNet-201\cite{b34}, and video classification models like I3D-Inception-v1, to benchmark event recognition. DenseNet-201 achieved the highest performance with an overall accuracy of 62.3\% in single-frame classification. The findings highlight the difficulties of recognizing events in a variety of environments and scales, noting that while models can identify specific events such as traffic congestion and smoke, they struggle with conditions such as night and snow scenes, indicating the need for improved attribute recognition and temporal cue exploitation in future research.

\begin{table}[htbp]
\centering
\caption{Summary of Research and Findings of UAV Datasets discussed}
\resizebox{\textwidth}{!}{%
\begin{tabular}{|p{2cm}|p{3cm}|p{3cm}|p{3.5cm}|p{3.5cm}|}
\hline
\textbf{Dataset} & \textbf{Dataset Details} & \textbf{Paper Findings} & \textbf{Limitations} & \textbf{Future Work} \\
\hline
RescueNet\cite{b1} & High-resolution images with pixel-level annotations for 10 classes, collected via UAVs after Hurricane Michael. & Attention-based and transformer-based methods performed best. Transfer learning from RescueNet to FloodNet improved segmentation. & Time-consuming annotation process and potential lack of comprehensive post-disaster elements. & Further evaluation across different disaster scenarios to enhance robustness. \\
\hline
UAV-Human\cite{b2} & 67,428 multi-modal video sequences for action recognition, pose estimation, person re-identification, and attribute recognition. & Highest action recognition accuracies with night-vision and IR videos. Pose estimation methods achieved mAP scores of 56.5 and 56.9. & Potential overfitting, lack of subject diversity, and constrained capturing conditions. & Increase sample size and diversity, and capture conditions to enhance model robustness and generalization. \\
\hline
AIDER\cite{b6} & 2565 manually gathered images of disaster events with augmentations. & Development of ERNet, achieving near state-of-the-art accuracy (90\%) and over 50 fps on a CPU platform. & Does not extensively discuss real-time implementation challenges, robustness in diverse conditions, or hyperparameter tuning. & Integrate ERNet with algorithms for detecting people and vehicles, use additional modalities like infrared cameras, and optimize the model for improved generalization and accuracy. \\
\hline
AU-AIR\cite{b3} & 32,823 labeled video frames with object annotations and flight data. & YOLOv3-tiny and MobileNetv2-SSD Lite for real-time object detection on UAVs showed potential for onboard computer applicability. & Focus on traffic surveillance may limit applicability to other scenarios, lacks advanced baselines for tasks like UAV navigation. & Enhance dataset diversity, incorporate more environmental contexts, and develop additional baselines leveraging sensor data for broader applications. \\
\hline
ERA\cite{b7} & 2,864 videos capturing events in a wide range of settings and sizes. & DenseNet-201 achieved the highest accuracy of 62.3\% in single-frame classification. & Dataset size, class imbalance, and challenge of distinguishing events from normal videos. & Focus on attribute recognition, temporal cue exploitation, and addressing challenging cases like human action recognition. \\
\hline
UAVid\cite{b4} & 30 video sequences featuring high-resolution 4K images with 8 labeled classes for semantic segmentation. & Multi-Scale-Dilation Net achieved an average IoU score of around 50\%. & Class imbalance, particularly in urban street scenes, potentially affecting model performance and generalization. & Balance method complexity with practical implementation, expand dataset size and object categories, address class imbalance, and explore other applications like object detection and tracking. \\
\hline
VRAI\cite{b5} & 137,613 images of 13,022 vehicles with detailed annotations captured by two UAVs. & Outperforms existing methods in vehicle ReID techniques using GANs and attention models. & Comparison scope, domain specificity, annotation complexity, scalability, and real-world deployment insights. & Explore transfer learning, enhance scalability, integrate advanced techniques, focus on real-world applications, and improve annotation strategies. \\
\hline
VERI-Wild\cite{b15} & Over 400,000 images of 40,671 vehicle IDs captured from a real CCTV camera system over one month. & FDA-Net outperforms existing methods, achieving highest Rank-1 and Rank-5 accuracies. & Potential biases due to urban district focus and dataset-specific adversarial scheme. & Explore more challenging real-world factors, generate comprehensive datasets, and leverage GANs to improve cross-view ReID performance. \\
\hline
\end{tabular}
}
\label{tab:dataset_summary_1}
\end{table}

\begin{table}[htbp]
\centering
\caption{Summary of Research and Findings of UAV Datasets discussed}
\resizebox{\textwidth}{!}{%
\begin{tabular}{|p{2cm}|p{3cm}|p{3cm}|p{3.5cm}|p{3.5cm}|}
\hline
\textbf{Dataset} & \textbf{Dataset Details} & \textbf{Paper Findings} & \textbf{Limitations} & \textbf{Future Work} \\
\hline
UAV-Assistant\cite{b9} & Data synthesis pipeline combining egocentric UAV views and exocentric user views with smooth silhouette loss. & Smooth silhouette loss enhances 3D pose estimation accuracy. & Lack of real-world data poses a challenge to generalizability, and determining optimal kernel size for smoothing filter. & Optimize parameters, explore additional loss functions, and validate approach in real-world scenarios. \\
\hline
KITE\cite{b10} & Focus on UAV control speech recognition with multimodal systems. & Recurrent neural networks (RNNs) for language modeling and visual cues integration. & Imperfect command-image associations, biases from semi-automatic methods for training data generation. & Address biases, enhance dataset generalizability, and explore other architectural decisions. \\
\hline
UAV-Gesture\cite{b11} & 119 high-definition video clips of 13 gestures for UAV navigation and command. & Annotates body joints and gesture classes in 37,151 frames using an extended version of VATIC. & Limited gesture set and non-expert actors may affect dataset quality. & Leverage dataset for gesture and action recognition in UAV control, expand and refine dataset for broader research applications. \\
\hline
DarkTrack 2021\cite{b12} & 110 annotated sequences totaling over 100,000 frames for low-light UAV tracking. & SCT demonstrated significant performance gains for nighttime UAV tracking. & Comparisons with daytime tracking scenarios needed to be improved. & Explore advanced transformer architectures, attention mechanisms, noise reduction strategies, and real-world validation. \\
\hline
UAVDark 135\cite{b13} & Over 125k manually annotated frames for dark tracking methods. & ADTrack demonstrates superiority in bright and dark conditions. & Lacks broader comparison with other state-of-the-art trackers. & Further research on real-time tracking algorithms, new image enhancement methods, multi-sensor fusion techniques, and hardware optimization strategies. \\
\hline
BioDrone\cite{b14} & 600 videos annotated and labeled at the frame level for single object tracking using bionic drone-based systems. & Comprehensive evaluation platform for robust vision research. & Focus on bionic UAVs may limit generalization, potential biases in annotations. & Improve tracking algorithms, address computational complexity and real-time performance. \\
\hline
FOR-Instance\cite{b8} & Five collections from around the world for individual tree segmentation from UAV-based laser scanning data. & Supports both instance and semantic segmentation, adaptable to deep learning frameworks. & Potential overfitting, lack of generalizability to other forest types, challenges with unclassified points. & Incorporate more data types, develop advanced deep learning architectures, study tree species classification, and conduct longitudinal studies on forest changes. \\
\hline
\end{tabular}
}
\label{tab:dataset_summary_2}
\end{table}

\subsection{UAVid}
Ye Lyu et al. introduced the UAVid\cite{b4} dataset in their paper which addresses the need for semantic segmentation in urban scenes from the perspective of UAVs. The UAVid dataset consists of 30 video sequences with 4K high-resolution images, which capture top and side views for improved object recognition and include 8 labeled classes. The paper highlights the challenges of large-scale variation, moving object recognition, and temporal consistency. The effectiveness of deep learning techniques, such as the Multi-Scale-Dilation net which is a novel technique proposed by the author, was evaluated and resulted in an average Intersection over Union\cite{b35} (IoU) score of approximately 50\%. Further enhancements were observed by employing spatial-temporal regularization methods like FSO\cite{b36} and 3D CRF\cite{b37}. The dataset's applicability extends to traffic monitoring, population density analysis, and urban greenery monitoring, showcasing its potential for diverse urban surveillance applications. The paper also discusses the dataset's class imbalance and suggests future expansions and optimizations to enhance its utility for semantic segmentation and other UAV-based tasks. 

\subsection{VRAI}
Peng Wang et al. introduced the VRAI\cite{b5} dataset, the largest vehicle re-identification (ReID) dataset with over 137,613 images of 13,022 vehicles. This UAV-based dataset includes annotations for unique IDs, color, vehicle type, attributes, and distinguishing features, capturing a wide range of view angles and poses from UAVs flying between 15m and 80m. The study devised an innovative vehicle ReID algorithm that utilizes weight matrices, weighted pooling, and comprehensive annotations to identify distinctive components. This algorithm surpasses both the baseline and the most advanced techniques currently available. The paper utilizes a comprehensive strategy to perform vehicle ReID using aerial images, showcasing its effectiveness through a range of experiments. Ablation study results demonstrate that the novel Multi-task + DP model, which integrates attribute classification and additional triplet loss on weighted features, exhibits superior performance compared to less complex models. The proposed method outperforms ground-based methods such as MGN\cite{b39}, RNN-HA\cite{b40}, and RAM\cite{b41}, because it can easily handle different view angles in UAV images. Weighted feature aggregation improves performance, as evidenced by the enhanced mean average precision (mAP) and cumulative match characteristic (CMC) metrics. Human performance evaluation highlights the algorithm's strength in fine-grained recognition, though humans still excel in detailed tasks. The study suggests further research to improve flexibility, scalability, and real-world application of the algorithm. 

\subsection{FOR-Instance}
For semantic and instance segmentation of individual trees, Stefano Puliti et al. presented the FOR-Instance\cite{b8} dataset in their paper "FOR-Instance: a UAV laser scanning benchmark dataset for semantic and instance segmentation of individual trees." This dataset fills a gap in the market for ML-ready datasets and standardized benchmarking infrastructure by offering publicly accessible annotated forest data for point cloud segmentation\cite{b38} tasks. The primary goal is to use data from unmanned aerial vehicle (UAV) laser scanning to precisely identify and separate individual trees. The dataset includes extensive annotations that are used for training and evaluation, and it is composed of five carefully chosen collections from different types of forests worldwide. In the context of deep learning, the dataset is divided into separate sets for the purpose of training and validation. In image segmentation research, rasterized canopy height models are utilized, along with either unprocessed point clouds or two-dimensional projections. The FOR-Instance dataset was found to be useful for studying and testing advanced segmentation methods. This highlights the significance of comprehending forest ecosystems and formulating sustainable management techniques. The standardization of the dataset in 3D forest scene segmentation research helps to address current methodological limitations, such as overfitting and lack of comparability.

\subsection{VERI-Wild}
Yihang Lou et al. presented the VERI-Wild\cite{b15} dataset, the largest vehicle ReID dataset to date, in their paper. Over 400,000 photos of 40,000 vehicle IDs are included in the dataset, which was collected over the course of a month in an urban district using 174 CCTV cameras. The dataset poses a formidable challenge for ReID algorithms due to its inclusion of diverse conditions such as varying backgrounds, lighting, obstructions, perspectives, weather, and vehicle types. The authors introduced FDA-Net, a novel technique for vehicle ReID, to enhance the model's ability to distinguish between different vehicles. FDA-Net combines a feature distance adversary network with a hard negative generator and embedding discriminator. After being tested on the VERI-Wild dataset and other established datasets, FDA-Net surpassed various standard methods, achieving higher accuracies in Rank-1 and Rank-5. This demonstrates the effectiveness of FDA-Net in vehicle ReID tasks. The method's ability to generate hard negatives significantly improved model performance, highlighting its potential for advancing vehicle ReID research in real-world scenarios. 

\subsection{UAV-Assistant}
G. Albanis and N. Zioulis et al. introduced the UAV-Assistant\cite{b9} (UAVA) dataset in their paper. The dataset was created using a data synthesis pipeline to generate realistic multimodal data, including exocentric and egocentric views from UAVs. The dataset can be utilized to train a model that can estimate the pose of an individual by incorporating a novel smooth silhouette loss in addition to a direct regression objective. The dataset can be used to train a model that can accurately determine the position of a person by incorporating a unique smooth silhouette loss along with a direct regression objective. It also uses differentiable rendering techniques to help the model learn from both real and fake data. The study highlights the critical role of tuning the kernel size for the smoothing filter to optimize model performance. The suggested smooth silhouette loss surpasses conventional silhouette loss functions by reducing discrepancies and enhancing the accuracy of 3D pose estimation. This approach specifically tackles the lack of available data for estimating the three-dimensional position and orientation of unmanned aerial vehicles (UAVs) in non-hostile environments. It is different from existing datasets that primarily focus on remote sensing or drones with malicious intent. The paper underscores the need for further research on rendering techniques, parameter optimization, and real-world validations to enhance the model's generalizability and robustness.

\subsection{KITE}
The KITE\cite{b10} dataset, created to improve speech recognition systems for UAV control, was presented by Dan Oneata and Horia Cucu in their paper. The KITE eval dataset is a comprehensive collection that includes 2,880 spoken commands, along with corresponding audio and images. It is specifically designed for UAV operations and covers a range of commands related to movement, camera usage, and specific scenarios. The authors employed time delay neural networks\cite{b42} (which is implemented in Kaldi\cite{b43}) and recurrent neural networks to perform language modeling. They initialized the models with out-of-domain datasets and subsequently fine-tuned them for UAV tasks. The study emphasizes the efficacy of customizing language models for UAV-specific instructions, showcasing substantial enhancements in speech recognition precision through domain adaptation. Future directions include grounding uttered commands in images for enhanced context understanding and improving the acoustic model's robustness to outdoor noises.

\subsection{UAV-Gesture}
A. Perera et al. introduced the UAV-Gesture\cite{b11} dataset, which addresses the lack of research on gesture-based UAV control in outdoor settings. This dataset aims to fill the existing research gap, as most studies in this field are focused on indoor environments. The dataset consists of 119 high-definition video clips, totaling 37,151 frames, captured in an outdoor setting using a 3DR Solo UAV and a GoPro Hero 4 Black camera. The dataset comprises annotations of 13 body joints and gesture classes for all frames, encompassing gestures appropriate for UAV navigation and command. The dataset was captured with variations in phase, orientation, and camera movement to augment realism. The authors employed an extended version of the VATIC\cite{b44} tool for annotation and utilized a Pose-based Convolutional Neural Network\cite{b45} (P-CNN) for gesture recognition. This approach resulted in a baseline accuracy of 91.9\%. This dataset facilitates extensive research in gesture recognition, action recognition, human pose recognition, and UAV control, showcasing its efficacy and potential for real-world applications.

\subsection{UAVDark135}
In their research Bowen Li et al. presented the UAVDark135\cite{b13} dataset and the ADTrack algorithm. Their work aimed to tackle the challenge of achieving reliable tracking of unmanned aerial vehicles (UAVs) under different lighting conditions. UAVDark135 is the inaugural benchmark specifically developed for tracking objects during nighttime. It consists of more than 125,000 frames that have been manually annotated, addressing a deficiency in current benchmarks. The paper details the ADTrack algorithm, a discriminative correlation filter-based tracker with illumination adaptive and anti-dark capabilities, utilizing image illuminance information and an image enhancer for real-time, all-day tracking. ADTrack performs better in both bright and dark environments, as evidenced by extensive testing on benchmarks such as UAV123@10fps\cite{b46}, DTB70\cite{b47}, and UAVDark135—achieving over 30 FPS on a single CPU. While effective, the paper recommends broader comparisons with other state-of-the-art trackers and future research on image enhancement, multi-sensor fusion, and UAV hardware optimization.

\subsection{DarkTrack2021}
Junjie Ye et al. presented the DarkTrack2021\cite{b12} dataset to tackle the difficulty of tracking unmanned aerial vehicles (UAVs) in low-light situations. The dataset consists of 110 annotated sequences containing more than 100,000 frames, providing a varied evaluation platform for tracking UAVs during nighttime. The researchers created an effective low-light enhancer called the Spatial-Channel Transformer (SCT), which combines a spatial-channel Transformer with a robust non-linear curve projection model to effectively enhance low-light images. The Spatial-Channel Attention Module (SCT) employs a technique that effectively combines global and local information, resulting in enhanced image quality by reducing noise and improving illumination in nighttime scenes. This study utilizes the proposed ADTrack algorithm together with 16 state-of-the-art handmade correlation filter (CF)-based trackers to evaluate their performance on tracking benchmarks UAV123@10fps, DTB70, and UAVDark135. The aim is to demonstrate the comprehensive robustness of the proposed ADTrack algorithm in all-day UAV tracking. Evaluations conducted on the public UAVDark135 and the new DarkTrack2021 benchmarks demonstrated that SCT exhibited superior performance compared to existing methods in tracking UAVs during nighttime. The practicality of the approach has been confirmed through real-world tests. The DarkTrack2021 dataset and SCT code are openly accessible on GitHub for additional research and experimentation.

\subsection{BioDrone}
Xin Zhao et al. presented the BioDrone\cite{b14} dataset. BioDrone is a pioneering visual benchmark for Single Object Tracking\cite{b49} (SOT) that utilizes bionic drones. It specifically tackles the difficulties associated with tracking small targets that undergo significant changes in appearance, which are common in flapping-wing UAVs. The dataset consists of 600 videos containing 304,209 frames that have been manually labeled. Additionally, there are automatically generated labels for ten challenge attributes at the frame level. The study presents a new baseline method, UAV-KT, optimized from KeepTrack\cite{b50}, and evaluates 20 SOT models, ranging from traditional approaches like KCF\cite{b51} to sophisticated models combining CNNs and SNNs. The results of comprehensive experiments demonstrate that UAV-KT outperforms other methods in handling challenging vision tasks with resilience. The paper emphasizes BioDrone's potential for advancing SOT algorithms and encourages future research to address remaining challenges, such as camera shake and dynamic visual environments.

\begin{figure}
\centering
\includegraphics[width=0.7\columnwidth]{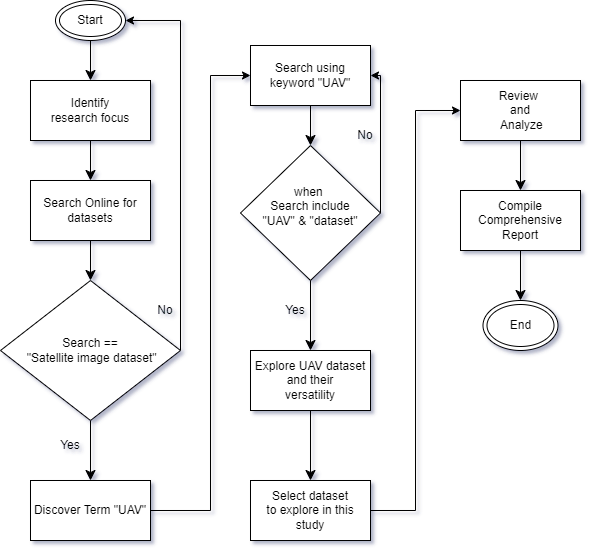}
\caption{Workflow of this Study}
\label{Methodology Flow Chart}
\end{figure}

\section{Methodology}
The term UAV (Unmanned Aerial Vehicle) encompasses a diverse range of applications, requiring a thorough investigation to examine and define the extensive utilization of UAV datasets. We aimed to comprehend how these datasets can be employed in different research and project scenarios. To accomplish this, we implemented an exhaustive search for UAV datasets, initially narrowing our focus to the keyword "satellite or drone image datasets". The initial search led to the identification of "UAV datasets". After acknowledging the potential of UAV datasets, we conducted further research in this field, identifying their diverse applications in object detection, tracking, and event detection, as well as semantic segmentation and single object tracking.

To gather relevant UAV datasets, we conducted systematic searches on the Internet, employing a range of keywords and search terms related to UAVs and their applications. We specifically looked for datasets that showed off the adaptability of UAVs, choosing those that researchers had proposed and used in other research contexts. This approach ensured that the datasets we included were novel and provided diverse examples of UAV applications.

We identified and collected 15 UAV image datasets for inclusion in our study. Our selection criteria focused on datasets that showcased a variety of use cases, including traffic systems (car identification, person identification, and surveillance systems), damage classification from disasters, and other object detection and segmentation tasks. Each dataset was thoroughly reviewed and analyzed to understand its characteristics, intended use, and underlying methodologies.

Our analysis involved a detailed examination of the datasets, resulting in the comprehensive report included in this paper. This report outlines the behavior, agenda, and applications of each dataset, providing insights into their respective fields of use. By presenting these findings, we aim to highlight the versatility and potential of UAV datasets in advancing various research domains. Figure \ref{Methodology Flow Chart} depicts the sequential process of our work.

\subsection{Search Terms}\label{AA}
We got the datasets we surveyed in this paper mostly from the website, \url{https://paperswithcode.com/}. Before we found this website we used various search terms to search for the UAV dataset and came across the website through the search process.
Example search strings:
\begin{itemize}
    \item ("unmanned aerial vehicle" OR UAV OR drone OR Satellite) AND ("dataset" OR "image dataset" OR "dataset papers")
    \item (UAV OR "unmanned aerial vehicle") AND ("disaster dataset" OR "traffic surveillance")
\end{itemize}

These search strings and keywords facilitated a broad yet focused search, enabling us to gather a diverse set of UAV datasets that demonstrate their wide-ranging applications and research potential.

\begin{table}[htbp]
\centering
\caption{Summary of Experimented Methods and Results on Different Datasets}
\begin{tabular}{|p{3cm}|p{5cm}|p{7cm}|}
\hline
\textbf{Dataset Name} & \textbf{Experimental Methods in Base Dataset publication} & \textbf{Analysis on Results} \\
\hline
RescueNet\cite{b1} & PSPNet, DeepLabv3+, Attention UNet, Segmenter\cite{b58} & Attention UNet achieved the best performance among all evaluated models. PSPNet showed better performance compared to DeepLabv3+ by using pyramid pooling. DeepLabv3+ provided moderate results, improving on the loss of boundary information. Segmenter showed varying results depending on the backbone (ViT-Tiny vs. ViT-Small), with heavier backbones achieving better results. \\
\hline
UAV-Human\cite{b2} & Guided Transformer I3D Network, Video Transformers, Full Model (Author's novel method) & Night-vision and IR videos outperformed previous findings in low-light conditions, achieving 28.72\% and 26.56\% accuracy, respectively. However, depth sequences face noise issues, and fisheye distortion impacts performance. In ablation studies, using KL Divergence Constraint resulted in 21.68\% accuracy, while employing guidance loss and Video Transformers yielded 21.49\% accuracy without RGB stream guidance. Overall, the full model had the highest accuracy among fisheye-based methods. \\
\hline
AIDER\cite{b6} & Novel networks (ERNet, SCFCNet, SCNet, baseNet), VGG16, ResNet50, MobileNet & The VGG16 model had the highest accuracy at 91.9\% but a low frame rate of 2, while consuming 59.3MB of memory. MobileNet had a high frame rate of 20 but lower accuracy at 88.5\%. Custom networks like ERNet and SCFCNet had good accuracy at 90.1\% and 87.7\% with high frame rates of 53 and 76, making them suitable for real-time UAV applications. \\
\hline
AU-AIR\cite{b3} & YOLOv3-tiny, MobileNetv2-SSD Lite & YOLOv3-tiny achieved higher mAP (38.2\%) and better FPS (22) compared to MobileNetv2-SSDLite (32.8\% mAP and 19 FPS), highlighting its better performance for real-time object detection tasks using UAVs. \\
\hline
ERA\cite{b7} & VGG-16, DenseNet-121, NASNet-L, C3D (C3D†, C3D‡)\cite{b59} & DenseNet-121 achieved the highest overall accuracy (62.3\%) among the models, followed by NASNet-L (60.2\%) and VGG-16 (51.9\%). The C3D models had the lowest accuracy (around 30\%). \\
\hline
UAVid\cite{b4} & FCN-8s\cite{b60}, Dilation Net, U-Net\cite{b61}, MS-Dilation Net & MS-Dilation Net achieved the highest mean IoU score of 57.3\% with pre-training and feature space optimization, demonstrating the best performance among the models evaluated. \\
\hline
VRAI\cite{b4} & MGN, RAM, RNN-HA, Ensemble methods (e.g. ID Classification Loss, Triplet + ID Loss), Novel methods (Multi-task, Multi-task + Discriminative Parts) & The multi-task model with discriminative parts achieved the highest mAP (78.63\%) and CMC-1 (80.30\%). The models using Triplet + ID Loss also showed high performance, particularly with Resnet-101 and Resnet-152 backbones. \\
\hline
FOR-instance\cite{b15} & None, as the paper is solely focused on constructing the dataset and explaining how to utilize it for model.  & \begin{center}N/A\end{center} \\
\hline
\end{tabular}
\label{tab:method_summary3}
\end{table}

\begin{table}[htbp]
\centering
\caption{Summary of Experimented Methods and Results on Different Datasets}
\begin{tabular}{|p{3cm}|p{5cm}|p{7cm}|}
\hline
\textbf{Dataset Name} & \textbf{Experimented Methods on Dataset} & \textbf{Analysis on results} \\
\hline
VERI-Wild\cite{b15} & GoogLeNet\cite{b62}, Triplet\cite{b63}, Softmax\cite{b64}, CCL\cite{b65}, HDC\cite{b66}, Unlabeled GAN\cite{b67}\cite{b68}, EN (Embedding Network with Triplet and Softmax Loss), FDA-Net $\ominus$ Att, FDA-Net & FDA-Net consistently outperforms the other models across different settings, achieving the highest mAP (35.11\%) and match rate (R=1 of 64.03\% for small dataset). The proposed FDA-Net model demonstrates its effectiveness in vehicle re-identification tasks. \\
\hline
UAV-Assistant (UAVA)\cite{b9} & Singleshotpose\cite{b69}, Direct, IoU based experimental methods (e.g. I0.1, I0.2, I0.1-0.4, G0.1, S0.1, S0.2), Generalized IoU based method (Gauss0.1)\cite{b70} & Gauss0.1 showed the best overall performance, particularly in the 6D Pose-5 and 6D Pose-10 metrics. Metrics such as NPE, OE, and CPE were used, with lower values indicating better performance and higher values for Acc5 and Acc10 indicating better performance. \\
\hline
KITE\cite{b10} & Baseline Systems (Unadapted System, Domain-Specific System), Domain Adaptation (Text-Only Adaptation, Rescoring), Multi-Modal Experiments (Text and Visual Information) & Domain adaptation and multi-modal approaches significantly improved the performance of speech recognition systems for UAV control. The Unadapted System had a WER of 56.2\%, while the Domain-Specific System achieved 11.7\%. \\
\hline
UAV-Gesture\cite{b11} & Pose-based CNN (P-CNN) & P-CNN achieved an overall accuracy of 91.9\% for gesture recognition. The dataset included 119 video clips, 37,151 annotated frames, and 13 gestures, providing a robust resource for gesture and action recognition research. \\
\hline
DarkTrack2021\cite{b12} & Novel ensembled method: SCT & The full implementation of SCT (Spatial-Channel Transformer) with all components enabled showed the highest improvement in tracking performance, with success rate and precision gains of 13.3\% and 15.4\%, respectively. \\
\hline
UAVDark135\cite{b13} & ADTrack, State of the art trackers (e.g. AutoTrack, SiamFC++, ARCF-HC, SiamRPN++) & ADTrack outperformed all other models in both bright and dark conditions, showing superior performance with the highest DP and AUC scores on the UAVDark135 dataset. \\
\hline
BioDrone\cite{b14} & KeepTrack, UAV-KT, Generic SOT Trackers & UAV-KT, designed for flapping-wing UAVs, showed a 5\% improvement over KeepTrack in precision, normalized precision, and success scores. Generic SOT Trackers were compared for robustness and performance across various conditions. \\
\hline
\end{tabular}
\label{tab:method_summary4}
\end{table}

\section{Data Diversity of UAV}
The advent of Unmanned Aerial Vehicles (UAVs) has opened new frontiers in data collection and analysis, transforming numerous fields with their versatile applications. The datasets generated by UAVs are diverse, encompassing various data types and serving multiple purposes. This section provides an overview of the various uses of UAV datasets, examines their diversity, and explores the methods applied to utilize these datasets in different studies.

\subsection{Overview of UAV Dataset Uses}
UAV datasets are pivotal in numerous domains, including disaster management, surveillance, agriculture, environmental monitoring, and human behavior analysis. The unique aerial perspectives provided by UAVs enable the collection of high-resolution imagery and videos, which can be used for mapping, monitoring, and analyzing different environments and activities.

\subsubsection{Disaster Management} 
UAV datasets are often used to figure out how much damage hurricanes, earthquakes, and floods have done. High-resolution images and videos captured by UAVs allow for precise mapping of affected areas and the identification of damaged infrastructure.

\subsubsection{Surveillance}
In urban and rural settings, UAV datasets support advanced surveillance activities. They facilitate the monitoring of traffic, detection of illegal activities, and overall urban planning by providing real-time, high-resolution aerial views.

\subsubsection{Agriculture}
UAV datasets help in monitoring crop health, assessing irrigation needs, and detecting pest infestations. Multispectral and hyperspectral imaging from UAVs enable detailed analysis of vegetation indices and soil properties.

\subsubsection{Environmental Monitoring}
UAVs are used to monitor forest health, wildlife, and water bodies. They provide data for studying ecological changes, tracking animal movements, and assessing the impacts of climate change.

\subsubsection{Human Behavior Analysis}
UAV datasets contribute to analyzing human activities and behaviors in public spaces. They are used for action recognition, pose estimation, and crowd monitoring, offering valuable insights for security and urban planning.

\subsection{Variability of UAV databases}
The diversity of UAV datasets lies in their varied data types, capture conditions, and application contexts. This diversity ensures that UAVs can address a wide range of tasks, each requiring specific data characteristics.

\subsubsection{Data Types}
UAV datasets include RGB images, infrared images, depth maps, and multispectral and hyperspectral images\cite{b52}. To capture complex scenarios for human behavior analysis, the UAV-Human dataset, for example, combines RGB videos, depth maps, infrared sequences, and skeleton data.

\subsubsection{Capture Conditions}
A variety of conditions, such as different times of day, weather, light (low light or varied lumination), and flight altitudes, are encountered when gathering UAV datasets. This variety makes sure that models that were trained on these datasets are strong and work well in a variety of settings.

\subsubsection{Application Contexts}
UAV datasets are tailored for specific applications. For example, visualizing data, object annotations, and flight data are used to address specific problems that come up when monitoring traffic from the air. Furthermore, the application of high-resolution images of the damage taken after the disaster, which enable accurate assessment of the damage.

\subsection{Methods Applied to the UAV Dataset} 
Various methods are applied to UAV datasets to extract valuable insights and solve specific problems. These methods include machine learning, computer vision techniques, and advanced data processing algorithms. In Table \ref{tab:method_summary3} and \ref{tab:method_summary4}, an overview of the methods used and the analysis of results are given to gain a better understanding.

\subsubsection{Machine Learning and Deep Learning}
Deep learning models, such as convolutional neural networks (CNNs)\cite{b53}, are widely used for tasks like object detection, segmentation, and classification. For example: 
\begin{itemize}
    \item The RescueNet dataset employs models like PSPNet, DeepLabv3+, and Attention UNet for semantic segmentation to assess disaster damage.  
    \item The UAVid Dataset presents deep learning baseline methods like Multi-Scale-Dilation net. The ERA dataset establishes a benchmark for event recognition in aerial videos by utilizing pre-existing deep learning models like the VGG models (VGG-16, VGG19)\cite{b28}, Inception-v3\cite{b55}, the ResNet models (ResNet-50, ResNet-101, and ResNet-152)\cite{b29}, MobileNet, the DenseNet models (DenseNet-121, DenseNet-169, DenseNet-201)\cite{b34}, and NASNet-L\cite{b56}. 
\end{itemize}
In the domain of deep learning, ensemble methods play a crucial role. They not only assess model performance but also boost accuracy while keeping the model’s equilibrium intact. Such as:
\begin{itemize}
    \item In VRAI dataset, they utilized ensemble techniques such as Triplet Loss, Contrastive Loss, ID Classification Loss, and Triplet + ID Loss, and introduced multi-task and multi-task + discriminative parts. These ensemble methods performed better than the state-of-the-art methods in their claim.
\end{itemize}

\subsubsection{Transfer Learning}
Transfer learning is used to leverage pre-trained models on UAV datasets, allowing for quicker and more efficient training. Like,
\begin{itemize}
    \item Pre-trained YOLOv3-Tiny and MobileNetv2-SSDLite models, for example, are used for real-time object detection in the AU-AIR\cite{b3} dataset.
\end{itemize}

\subsubsection{Event Recognition}
Unmanned Aerial Vehicles (UAVs) have proven to be highly proficient in the field of event recognition and have gained significant popularity in this domain. Like for example:
\begin{itemize}
    \item The ERA dataset has been subjected to various methods for event recognition in aerial videos, including DenseNet-201 and Inception-v3. These methods have demonstrated notable accuracy in identifying dynamic events from UAV footage. 
    \item The BioDrone dataset assesses single object tracking (SOT) models and investigates new optimization approaches for the cutting-edge KeepTrack method for robust vision, which is presented by flapping-wing unmanned aerial vehicles\cite{b14}.
\end{itemize}

\subsubsection{Multimodal Analysis}
Combining data from multiple sensors enhances the analysis capabilities of UAV datasets. The multimodal approach of the UAV-Human dataset, which combines RGB, infrared, and depth data, makes a thorough analysis of human behavior possible.

\subsubsection{Creative Algorithms}
New algorithms are created to tackle particular problems in the analysis of data from unmanned aerial vehicles. For example: 
\begin{itemize}
    \item The UAV-Gesture\cite{b11} dataset employs advanced gesture recognition algorithms to enable UAV navigation and control based on human gestures. 
    \item The UAVDark135\cite{b13} makes use of ADTrack, a tracker that adapts to varying lighting conditions and makes use of discriminative correlation filters. It also has anti-dark capabilities.
    \item To address the issue of fisheye video distortions, the authors of the UAV-Human\cite{b2} dataset suggest a fisheye-based action recognition method that uses flat RGB videos as guidance. 
    \item To classify disaster events from an unmanned aerial vehicle (UAV), the authors of the AIDER\cite{b6} dataset have created a lightweight convolutional neural network (CNN) architecture that they have named ERNet. 
    \item VERI-Wild\cite{b15} introduces FDA-Net, a novel method for vehicle identification. It includes an embedding discriminator and a feature distance adversary network to enhance the model's capacity to differentiate between various automobiles.
\end{itemize}

\subsubsection{Managing Diverse Conditions}
Various environmental conditions, such as different lighting, weather, and occlusions, present challenges that are often addressed by methodologies. Like, DarkTrack2021 used the low-light enhancer-based method SCT to handle performance in low-light conditions.

The diversity of UAV datasets is a cornerstone of their utility, enabling a wide array of applications across different fields. From disaster management to human behavior analysis, the rich variety of data types, capture conditions, and application contexts ensures that UAV datasets can meet the specific needs of each task. The application of advanced methods, including deep learning, transfer learning, and multimodal analysis, further enhances the value derived from these datasets, pushing the boundaries of what UAVs can achieve in research and practical applications.

\begin{table*}[htbp]
\centering
\caption{Summary of Methods Employed on Uav Datasets and Their Benefits}
\resizebox{\textwidth}{!}
{%
\begin{tabular}{|p{4cm}|p{3cm}|p{8cm}|}
\hline
\textbf{Employed Method} & \textbf{Name of the Dataset} & \textbf{Benefit from the Use of Method} \\
\hline
Attention UNet28, ViT-Tiny, ViT-Small & RescueNet\cite{b1} & Improved disaster response strategies, enhanced model performance in segmentation tasks through transfer learning \\
\hline
Fisheye-based action recognition approach, HigherHRNet, AlphaPose & UAV-Human\cite{b2} & Robust models for human behavior understanding \\
\hline
ERNet & AIDER\cite{b6} & High performance with minimal memory requirements, suitable for real-time aerial image classification \\
\hline
YOLOv3-Tiny, MobileNetv2-SSDLite & AU-AIR\cite{b3} & Real-time object detection on UAVs, bridging the gap between computer vision and robotics \\
\hline
DenseNet-201, I3D-Inception-v1, TRN-Inception-v3 & ERA\cite{b7} & High performance in single-frame and video classification tasks \\
\hline
Multi-Scale-Dilation net, FSO, 3D CRF & UAVid\cite{b4} & Enhanced semantic segmentation performance in urban scenes, addressing large-scale variation and moving object recognition \\
\hline
Convolutional and connection layers, weight matrices, weighted pooling & VRAI\cite{b5} & Superior vehicle re-identification performance \\
\hline
Aggregating tree-wise F1 scores, weighting coefficients for averaging F1 scores & FOR-instance\cite{b8} & Improved methods for individual tree segmentation, crucial for understanding forest ecosystems \\
\hline
FDA-Net (Feature Distance Adversary Network) & VERI-Wild\cite{b15} & Enhanced discriminative capability in vehicle re-identification tasks \\
\hline
Smooth silhouette loss & UAV-Assistant (UAVA)\cite{b9} & Improved performance in 3D pose estimation tasks \\
\hline
Time delay neural network, domain adaptation techniques & KITE\cite{b10} & Enhanced UAV command recognition systems through visual context and domain adaptation \\
\hline
Pose-based Convolutional Neural Network (P-CNN) & UAV-Gesture\cite{b11} & High accuracy in gesture recognition for UAV control \\
\hline
Spatial-Channel Transformer, curve projection model & DarkTrack2021\cite{b12} & Improved nighttime UAV tracking accuracy by enhancing low-light images \\
\hline
Illumination adaptive, anti-dark capabilities, efficient image enhancer & UAVDark135\cite{b13} & Superior performance in all-day aerial object tracking, adaptability to different light conditions \\
\hline
KeepTrack-optimized UAV-KT & BioDrone\cite{b14} & Addresses challenges in tracking tiny targets with drastic appearance changes, providing a robust benchmark for vision research \\
\hline
\end{tabular}%
}
\label{tab:methods_summary}
\end{table*}

\section{The Potential of Computer Vision Research in UAV Datasets}

\begin{figure}
\centering
\includegraphics[width=0.7\columnwidth]{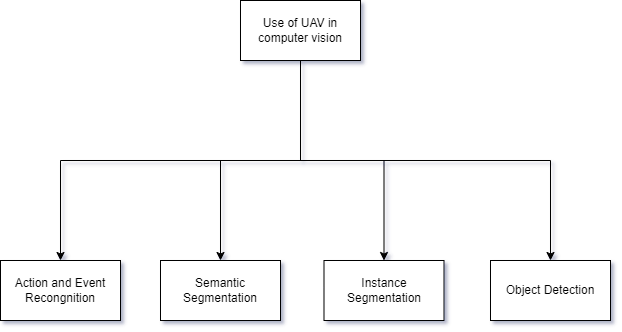}
\caption{Diverse Applications of UAV Datasets in Computer Vision Research}
\label{Diverse Applications of UAV  Flowchart}
\end{figure}

Unmanned Aerial Vehicles (UAVs) have greatly expanded the fields of computer vision research. UAV datasets offer unique and flexible data that is used in a range of computer vision tasks, from recognizing actions to finding objects. This section explores how UAV datasets are advancing computer vision research, contributing to various tasks from action recognition to object detection, as illustrated in Figure \ref{Diverse Applications of UAV  Flowchart}, which highlights the diverse applications and the development of new methods centered around these datasets. 

\subsection{Leveraging UAV Datasets for Computer Vision Applications}

Human behavior analysis, emergency response, tracking at night, surveillance, and many other uses can be done with UAV datasets in computer vision. These are some of the areas where UAV datasets are used, along with an example of how to describe a dataset based on the datasets we talked about in our research paper.

\subsubsection{Human Behavior Understanding and Gesture Recognition}
The UAV-Human platform is essential for utilizing UAVs to study human behavior, including a range of conditions and perspectives for pose estimation and action recognition. This dataset contains multi-modal information, including skeleton, RGB, infrared, and night vision modalities.
Essential for UAV control and gesture identification, UAV-Gesture contains 119 high-definition video clips with 13 gestures for command and navigation that are marked with body joints and gesture classes. Because this dataset was captured outside, it has more practical UAV control applications because of the variations in phase, orientation, and body shape.

\subsubsection{Emergency Response and Disaster Management}
RescueNet provides detailed pixel-level annotations and high-resolution images for 10 classes, including buildings, roads, pools, and trees. It is designed for post-disaster damage assessment using UAV imagery. It supports semantic segmentation using state-of-the-art models, enhancing natural disaster response and recovery strategies.
AIDER focuses on classifying disaster events, utilizing images of traffic accidents, building collapses, fires, and floods to support real-time disaster management applications by training convolutional neural networks (CNNs).

\subsubsection{Traffic Surveillance and Vehicle Re-Identification}
In traffic surveillance, AU-AIR prioritizes real-time performance and offers annotations for a variety of object categories, including cars, buses, and pedestrians. It bridges the gap between computer vision and robotics by offering multi-modal sensor data for advanced research in data fusion applications.
VRAI is the largest UAV-based vehicle re-identification dataset, containing over 137,613 images of 13,022 vehicles with annotations for unique IDs, color, vehicle type, attributes, and discriminative parts. It supports vehicle ReID tasks with diverse scenarios and advanced algorithms.
VERI-Wild, which contains over 400,000 photos of 40,000 vehicles taken by 174 CCTV cameras in various urban settings, is essential for research on vehicle re-identification. It uses techniques like FDA-Net to improve ReID accuracy by addressing variations in backgrounds, illumination, occlusion, and viewpoints.

\subsubsection{Event Recognition and Video Understanding}
For training models in event recognition in UAV videos, ERA contains 2,864 labeled video snippets for 24 event classes and 1 normal class that were gathered from YouTube. This dataset captures dynamic events in various conditions, supporting temporal event localization and video retrieval tasks.

\subsubsection{Nighttime tracking and low-light conditions}
Including 110 annotated sequences with over 100,000 frames, DarkTrack2021 is crucial for improving UAV tracking at night. By employing spatial-channel transformers (SCT) and non-linear curve projection models, it improves the quality of low-light images and offers a thorough assessment framework. 
The UAVDark135 dataset and the ADTrack algorithm are designed for all-day aerial tracking. ADTrack performs well in low light and adjusts to various lighting conditions thanks to its discriminative correlation filter foundation. More than 125,000 frames, specially annotated for low-light tracking scenarios, are included in the UAVDark135 dataset.

\subsubsection{Object Tracking and Robust Vision}
With 600 videos and 304,209 manually labeled frames, BioDrone is a benchmark for single object tracking with bionic drones. It captures challenges such as camera shake and drastic appearance changes, supporting robust vision analyses and evaluations of various single object tracking algorithms.

\subsubsection{Urban Scene Segmentation and Forestry Analysis}
UAVid provides annotations for eight classes and 30 high-resolution video sequences in 4K resolution to address segmentation challenges in urban scenes. It uses models such as Multi-Scale-Dilation net to support tasks like population density analysis and traffic monitoring.
FOR-instance provides UAV-based laser scanning data for tree instance segmentation and is intended for use in point cloud segmentation in forestry. It facilitates benchmarking and method development by supporting both instance and semantic segmentation.

\subsubsection{Multimodal Data Synthesis and UAV Control}
UAV-Assistant facilitates monocular pose estimation by introducing a multimodal dataset featuring exocentric and egocentric views. It enhances 3D pose estimation tasks with novel smooth silhouette loss function and differentiable rendering techniques.
KITE incorporates spoken commands, audio, and images to enhance UAV control systems. It includes commands recorded by 16 speakers, supporting movement, camera-related, and scenario-specific commands with multi-modal approaches.

Together, these datasets improve a wide range of computer vision applications, including robust vision in difficult conditions, real-time traffic surveillance, emergency response, and human behavior analysis.

\subsection{Development of Novel Methods Using UAV Datasets}

UAV datasets have spurred the development of innovative methods in computer vision. As an example, the Guided Transformer I3D framework, which addresses distortions through unbounded transformations guided by flat RGB videos, was developed using the UAV-Human dataset. This framework enhances action recognition performance in fisheye videos. This approach is a prime example of how UAV datasets drive the creation of specialized algorithms to address particular difficulties brought about by aerial viewpoints.

The DarkTrack2021 benchmark introduces a Spatial-Channel Transformer (SCT) for enhancing low-light images in nighttime UAV tracking. Meanwhile, Bowen Li and team present the UAVDark135 dataset and the ADTrack algorithm for all-day aerial object tracking. ADTrack, equipped with adaptive illumination and anti-dark capabilities, outperforms other trackers in both well-lit and dark conditions. It processes over 30 frames per second on a single CPU, ensuring efficient tracking under various lighting conditions. The study emphasizes how crucial image illuminance data is and suggests a useful image enhancer to improve tracking performance in all-day situations.

For emergency response applications, the AIDER dataset has facilitated the development of ERNet, a lightweight CNN architecture optimized for embedded platforms. ERNet's architecture, which incorporates downsampling at an early stage and efficient convolutional layers, allows for real-time classification of aerial images on low-power devices. This showcases the practical use of UAV datasets in disaster management. 

The VERI-Wild dataset introduces a novel approach called FDA-Net for vehicle reidentification. This method utilizes a unique type of network to generate difficult negative examples in the feature space. On the other hand, the VRAI dataset has developed a specialized vehicle ReID algorithm that leverages detailed annotation information to explicitly identify unique parts for each vehicle instance in object detection. 

Ultimately, UAV datasets are essential in the field of computer vision research, providing distinct data that is invaluable for a diverse array of applications. They allow for the development of novel methods tailored to the specific challenges and opportunities presented by UAV technology, accelerating progress in areas such as human behavior analysis, emergency response, and nighttime tracking.

\section{Constraints of UAVs}
While Unmanned Aerial Vehicles (UAVs) have significantly advanced data collection and analysis in numerous fields, they are not without limitations, particularly concerning the datasets they generate. This section delves into the primary constraints associated with UAV datasets, emphasizing their impact on the field and suggesting areas for improvement.

\subsection{Data Quality and Consistency}
One of the most pressing limitations of UAV datasets is the inconsistency in data quality. Weather, time of day, and UAV stability are just a few variables that can affect the quality of data that UAVs collect. Such as, datasets collected during poor weather conditions or at night may need more visibility and increased noise, complicating subsequent analysis and model training. Even with advancements like low-light image enhancers and specialized algorithms for nighttime tracking, these solutions often need improvement and require further refinement to match the reliability of daytime data.

\subsection{Limited Scope and Diversity}
UAV datasets often need more diversity in terms of geographic locations, environmental conditions, and the variety of captured objects. Many existing datasets, such as AU-AIR and ERA, focus heavily on specific scenarios like urban traffic surveillance or disaster response, which limits their generalizability to other contexts. Additionally, datasets such as UAV-Human and UAVDark135 tend to feature limited subject diversity and controlled environments, which may not accurately represent real-world conditions. This lack of diversity can lead to models that perform well in specific conditions but struggle in untested environments.

\subsection{Annotation Challenges}
The process of annotating UAV datasets is often time-consuming and labor-intensive. High-resolution images and videos captured by UAVs require detailed, pixel-level annotations, which are essential for tasks like semantic segmentation and object detection. This is clearly seen in datasets such as RescueNet and FOR-Instance, where the annotation process is recognized as a major bottleneck. The intensive labor required for comprehensive annotation limits the availability of large, well-labeled datasets, which are crucial for training robust machine learning models.

\subsection{Computational and Storage Demands}
The high resolution and large volume of data generated by UAVs pose significant computational and storage challenges. Processing and analyzing large-scale UAV datasets demand substantial computational resources and advanced hardware, which may only be readily available to some researchers. For example, the dense and high-resolution images in datasets like UAVid and BioDrone require extensive processing power for effective utilization. Additionally, the storage of such vast amounts of data can be impractical for some institutions, hindering widespread access and collaboration.

\subsection{Integration with Other Data Sources}
Another limitation is the integration of UAV datasets with other data sources. While multimodal datasets that combine UAV data with other sensor inputs (such as satellite imagery, GPS data, and environmental sensors) provide richer insights, they also introduce complexity in data alignment and fusion. The AU-AIR dataset, which includes visual data along with GPS coordinates and IMU data, exemplifies the potential and challenges of such integration. Ensuring the synchronized and accurate fusion of data from multiple sources remains a technical hurdle that needs addressing.

\subsection{Real-Time Data Processing}
The ability to process and analyze UAV data in real-time is critical for applications like disaster response and surveillance. However, achieving real-time processing with high accuracy is challenging due to the aforementioned computational demands. Models such as those evaluated in the DarkTrack2021 and UAVDark135 datasets show promise but often require optimization to balance speed and accuracy effectively. Real-time processing also necessitates robust algorithms capable of handling dynamic environments and changing conditions without significant delays.

\subsection{Ethical and Legal Considerations}
Finally, the use of UAVs and their datasets is subject to various ethical and legal considerations. Issues such as privacy, data security, and regulatory compliance must be addressed to ensure responsible and lawful use of UAV technology. These considerations can limit the scope of data collection and usage, particularly in populated areas or sensitive environments, thereby constraining the availability and applicability of UAV datasets.

Despite the transformative potential of UAV datasets across various disciplines, their limitations must be acknowledged and addressed to maximize their utility. Improving data quality, enhancing dataset diversity, streamlining annotation processes, and overcoming computational and storage challenges are essential steps. Additionally, integrating UAV data with other sources, advancing real-time processing capabilities, and adhering to ethical and legal standards will ensure that UAV datasets can be effectively leveraged for future research and applications. By tackling these limitations, the field can fully harness the power of UAV technology to drive innovation and deepen our understanding of complex, dynamic environments from an aerial perspective.

\section{Prospects for Future UAV Research}
Future studies on UAV datasets need to focus on a few crucial areas to improve their usefulness and cross-domain applicability as the field grows. The following suggestions highlight the crucial paths for creating UAV datasets and maximizing their potential for future innovations.

\subsection{Enhancing Dataset Diversity and Representativeness}
Further investigations ought to concentrate on generating more varied and representative UAV datasets. This involves capturing data in a wider range of environments, weather conditions, and geographic locations to ensure models trained on these datasets are robust and generalizable. To obtain comprehensive data for tasks like environmental monitoring, urban planning, and disaster response, datasets can be expanded to include a variety of urban, rural, and natural settings.

\subsection{Incorporating Multimodal Data Integration}
Integrating multiple data modalities, such as thermal, infrared, LiDAR\cite{b57}, and hyperspectral\cite{b52} imagery, can significantly enrich UAV datasets. In the future, these data types should be combined to create multimodal datasets that provide a more comprehensive view of the scenes that were recorded. This integration can improve the accuracy of applications such as vegetation analysis, search and rescue operations, and wildlife monitoring.

\subsection{Advancing Real-Time Data Processing and Transmission}
For applications like emergency response and traffic monitoring that demand quick analysis and decision-making, developing techniques for real-time data processing and transmission is essential. Future research should focus on optimizing data compression, transmission protocols, and edge computing techniques to enable swift and efficient data handling directly on UAVs.

\subsection{Improving Annotation Quality and Efficiency}
High-quality annotations are vital for the effectiveness of UAV datasets in training machine learning models. Future studies should investigate automated and semi-automated annotation tools that leverage AI to reduce manual labor and improve annotation accuracy. Additionally, crowdsourcing and collaborative platforms can be utilized to gather diverse annotations, further enhancing dataset quality.

\subsection{Addressing Ethical and Privacy Concerns}
As UAVs become more prevalent, addressing ethical and privacy issues becomes increasingly important. Guidelines and frameworks for the ethical use of UAV data should be established by future research, especially for applications involving surveillance and monitoring. It is important to focus on creating methods that protect privacy and collect data in a way that respects regulations and earns the trust of the public.

\subsection{Expanding Application-Specific Datasets}
The creation of customized datasets for specific uses can effectively boost new ideas in certain areas. For instance, datasets focused on agricultural monitoring, wildlife tracking, or infrastructure inspection can provide domain-specific insights and improve the precision of related models. To address the specific needs of various industries, future research should give priority to developing such targeted datasets.

\subsection{Enhancing Interoperability and Standardization}
Standardizing data formats and annotation protocols across UAV datasets can make it easier for researchers and developers to use and make the datasets more interoperable. Future efforts should aim to establish common standards and benchmarks, enabling the seamless integration of datasets from various sources and promoting collaborative research efforts.

\subsection{Utilizing Advanced Machine Learning Techniques}
The application of cutting-edge machine learning techniques, such as deep learning and reinforcement learning, to UAV datasets holds immense potential for advancing UAV capabilities. Future research should explore innovative algorithms and models that can leverage the rich data provided by UAVs to achieve breakthroughs in areas like autonomous navigation, object detection, and environmental monitoring.

\subsection{Leveraging Advanced Machine Learning Techniques}
Longitudinal studies that collect UAV data over long periods of time can give us useful information about how things change over time in different settings. Future research should emphasize continuous data collection efforts to monitor changes in ecosystems, urban developments, and disaster-prone areas, enabling more informed and proactive decision-making.
\subsection{Fostering Collaborative Research and Open Data Initiatives}
Encouraging collaboration among researchers, institutions, and industries can accelerate advancements in UAV datasets. Open data initiatives that make UAV datasets public should be supported by future research. These initiatives will encourage innovation and allow a wider range of researchers to contribute to and use these resources.

By addressing these future research directions, the field of UAV datasets can continue to evolve, offering increasingly sophisticated tools and insights that drive progress across multiple domains. UAV datasets are still being improved and added to, which is very important for getting the most out of UAV technology and making room for new discoveries and uses.

\begin{table}[htbp]
    \centering
    \caption{Performance Metrics and Results for Different Datasets and Methods}
    \resizebox{\textwidth}{!}{%
    \begin{tabular}{|C{3cm}|C{3cm}|C{3cm}|C{6cm}|}
        \hline
        \textbf{Dataset Name} & \textbf{Reference} & \textbf{Methods} & \textbf{Performance} \\ 
        
        \hline
        \multirow{6}{*}{AU-AIR\cite{b3}} 
            & \multirow{3}{*}{\cite{b71}} 
                & YOLOv3 
                    & \begin{tabular}{c|c}
                            mAP & Speed(FPS) \\ 
                            \hline
                            59.83 & 29 \\ 
                        \end{tabular} \\
                \cline{3-4}       
                & & YOLOv4 
                    & \begin{tabular}{c|c}
                            mAP & Speed(FPS) \\ 
                            \hline
                            67.35 & 24 \\ 
                        \end{tabular} \\ 
                \cline{3-4}
                & & RSSD-TA-LSTM-GID 
                    & \begin{tabular}{c|c}
                            mAP & Speed(FPS) \\ 
                            \hline
                            71.68 & 23 \\ 
                        \end{tabular} \\ 
            \cline{2-4}

            & \multirow{3}{*}{\cite{b72}} 
                & res2net50 
                    & \begin{tabular}{c|c}
                            mAP & Speed(FPS) \\ 
                            \hline
                            88.93 & 45.73 \\ 
                        \end{tabular} \\
                \cline{3-4}       
                & & rs2net101 
                    & \begin{tabular}{c|c}
                            mAP & Speed(FPS) \\ 
                            \hline
                            90.52 & 7.21 \\ 
                        \end{tabular} \\ 
                \cline{3-4}
                & & hourglass-104 
                    & \begin{tabular}{c|c}
                            mAP & Speed(FPS) \\ 
                            \hline
                            91.62 & 7.19 \\ 
                        \end{tabular} \\ 
            \cline{2-4}

            & \multirow{3}{*}{\cite{b73}} 
                & RetinaNet 
                    & \begin{tabular}{c|c}
                            Voting Strategy & mAP(\%) \\ 
                            \hline
                            Unanimous & 6.63 \\
                        \end{tabular} \\
                \cline{3-4}       
                & & YOLO + RetinaNet
                    & \begin{tabular}{c|c}
                            Voting Strategy & mAP(\%) \\ 
                            \hline
                            Consensus &  3.69 \\ 
                        \end{tabular} \\ 
                \cline{3-4}
                & & RetinaNet + SSD
                    & \begin{tabular}{c|c}
                            Voting Strategy & mAP(\%) \\ 
                            \hline
                            Consensus & 4.03 \\ 
                        \end{tabular} \\
                \cline{2-4}

                & \multirow{3}{*}{\cite{b74}} 
                & Faster R-CNN 
                    & \begin{tabular}{c}
                            mAP(\%)  \\ 
                            \hline
                            13.77  \\ 
                        \end{tabular} \\
                \cline{3-4}       
                & & SSD 
                    & \begin{tabular}{c}
                            mAP(\%) \\ 
                            \hline
                            9.1  \\ 
                        \end{tabular} \\ 
                \cline{3-4}
                & & YOLOv3 
                    & \begin{tabular}{c}
                            mAP(\%)  \\ 
                            \hline
                            13.33 \\ 
                        \end{tabular} \\
                \cline{3-4}
                & & YOLOv4  
                    & \begin{tabular}{c}
                            mAP(\%)  \\ 
                            \hline
                            25.94 \\ 
                        \end{tabular} \\
            \cline{1-4}

            \hline
            \multirow{6}{*}{FOR-instance\cite{b8}} 
                & \multirow{3}{*}{\cite{b80}} 
                    & PointNet 
                        & \begin{tabular}{c|c}
                                mIoU & micro F1 \\ 
                                \hline
                                35.65 & 52.56 \\ 
                            \end{tabular} \\
                    \cline{3-4}       
                    & & PointNet++
                        & \begin{tabular}{c|c}
                                mIoU & micro F1 \\ 
                                \hline
                                 33.00 & 49.57 \\ 
                            \end{tabular} \\ 
                    \cline{3-4}
                    & & Point Transformers
                        & \begin{tabular}{c|c}
                                mIoU & micro F1 \\  
                                \hline
                                22.97 & 37.13 \\ 
                            \end{tabular} \\ 
                \cline{2-4}
    
                & \multirow{3}{*}{\cite{b81}} 
                    & HFC (on CULS plot\tablefootnote{"Plot" here refers to the forest types that were looked at in the FOR-instance dataset release.}) 
                        & \begin{tabular}{c|c|c}
                                Precision & Recall & F1 score \\ 
                                \hline
                                0.89 & 0.8 & 0.84 \\ 
                            \end{tabular} \\
                    \cline{3-4}       
                    & & HFC (on NIBIO plot) 
                        & \begin{tabular}{c|c|c}
                                Precision & Recall & F1 score \\ 
                                \hline
                                0.89 & 0.85 & 0.87\\ 
                            \end{tabular} \\
                                        \cline{3-4}       
                    & & HFC (on NIBIO2 plot) 
                        & \begin{tabular}{c|c|c}
                                Precision & Recall & F1 score \\ 
                                \hline
                                0.85 & 0.85 & 0.85\\ 
                            \end{tabular} \\
                                                \cline{3-4}       
                    & & HFC (on SCION plot) 
                        & \begin{tabular}{c|c|c}
                                Precision & Recall & F1 score \\ 
                                \hline
                                0.95 & 0.90 & 0.92\\ 
                            \end{tabular} \\
                                                \cline{3-4}       
                    & & HFC (on RMIT plot) 
                        & \begin{tabular}{c|c|c}
                                Precision & Recall & F1 score \\ 
                                \hline
                                0.89 & 0.85 & 0.87\\ 
                            \end{tabular} \\
                                                \cline{3-4}       
                    & & HFC (on TUWEIN plot)
                        & \begin{tabular}{c|c|c}
                                Precision & Recall & F1 score \\ 
                                \hline
                                0.84 & 0.80 & 0.82\\ 
                            \end{tabular} \\
                \cline{1-4}

            \hline
            \multirow{6}{*}{UAV-Assistant\cite{b9}} 
                & \multirow{3}{*}{\cite{b88}} 
                    & BPnP\cite{b89} 
                        & \begin{tabular}{c|c}
                                ACC2 & ACC5 \\ 
                                \hline
                                95.2 & 98.36 \\
                                \hline
                                55.31 & 85.34 \\
                            \end{tabular} \\
                    \cline{3-4}       
                    & & HigherHRNet\cite{b90}
                        & \begin{tabular}{c|c}
                                ACC2 & ACC5 \\ 
                                \hline
                                89.92 & 97.75 \\
                            \end{tabular} \\
                    \cline{3-4}       
                    & & HRNet\cite{b91}
                        & \begin{tabular}{c|c}
                                ACC2 & ACC5 \\ 
                                \hline
                                90.75 & 98.04 \\
                            \end{tabular} \\
                \cline{1-4}


    \end{tabular}
    }
    \label{tab:performance}
\end{table}

\begin{table}[htbp]
    \centering
    \caption{Performance Metrics and Results for Different Datasets and Methods}
    \resizebox{\textwidth}{!}{%
    \begin{tabular}{|c|c|c|c|}
        \hline
        \textbf{Dataset Name} & \textbf{Reference} & \textbf{Methods} & \textbf{Performance} \\ 
            \hline
            \multirow{6}{*}{AIDER\cite{b6}} 
                & \multirow{3}{*}{\cite{b75}} 
                    & EmergencyNet 
                        & \begin{tabular}{c|c}
                                memory(MB) & F1 Score(\%) \\ 
                                \hline
                                0.368 & 95.7 \\ 
                            \end{tabular} \\
                    \cline{3-4}       
                    & & VGG16
                        & \begin{tabular}{c|c}
                                memory(mB) & F1 Score(\%) \\ 
                                \hline
                                 59.39 & 96.4 \\ 
                            \end{tabular} \\ 
                    \cline{3-4}
                    & & ResNet50 
                        & \begin{tabular}{c|c}
                                memory(MB) & F1 Score(\%) \\ 
                                \hline
                                96.4 & 96.1 \\ 
                            \end{tabular} \\ 
                \cline{2-4}
    
                & \multirow{3}{*}{\cite{b76}} 
                    & AISCC-DE2MS 
                        & \begin{tabular}{c|c}
                                MSE & PSNR \\ 
                                \hline
                                0.042 & 61.898 \\ 
                            \end{tabular} \\
                    \cline{3-4}       
                    & & Genetic Algorithm 
                        & \begin{tabular}{c|c}
                                MSE & PSNR \\ 
                                \hline
                                0.06 & 60.349 \\ 
                            \end{tabular} \\ 
                    \cline{3-4}
                    & & Cat Swarm Algorithm 
                        & \begin{tabular}{c|c}
                                MSE & PSNR \\ 
                                \hline
                                0.12 & 57.339 \\ 
                            \end{tabular} \\
                    \cline{3-4}
                    & & Artificial Bee Colony Algorithm 
                        & \begin{tabular}{c|c}
                                MSE & PSNR \\ 
                                \hline
                                0.165 & 55.956 \\ 
                            \end{tabular} \\
                \cline{1-4}
            \hline
            \multirow{6}{*}{DarkTrack2021\cite{b12}} 
                & \multirow{3}{*}{\cite{b77}} 
                    & SAM-DA-Track 
                        & \begin{tabular}{c|c|c}
                                AUC & Precision (normalized) & Precision \\ 
                                \hline
                                0.451 & 0.524 & 0.593 \\ 
                            \end{tabular} \\
                    \cline{3-4}       
                    & & UDAT
                        & \begin{tabular}{c|c|c}
                                AUC & Precision (normalized) & Precision \\ 
                                \hline
                                 0.421 & 0.499 & 0.570 \\ 
                            \end{tabular} \\ 
                    \cline{3-4}
                    & & SiamBAN
                        & \begin{tabular}{c|c|c}
                                AUC & Precision (normalized) & Precision \\ 
                                \hline
                                0.422 & 0.491 & 0.566 \\ 
                            \end{tabular} \\ 
                \cline{2-4}
    
                & \multirow{3}{*}{\cite{b78}} 
                    & SiamAPN 
                        & \begin{tabular}{c|c|c}
                                DP\tablefootnote{DP: Dynamic Precision} & NDP\tablefootnote{NDP: Normalized Dynamic Precision} & AUC \\ 
                                \hline
                                0.43 & 0.389 & 0.446 \\ 
                            \end{tabular} \\
                    \cline{3-4}       
                    & & SiamAPN++ 
                        & \begin{tabular}{c|c|c}
                                DP & NDP & AUC \\ 
                                \hline
                                0.494 & 0.446 & 0.375\\ 
                            \end{tabular} \\ 
                \cline{1-4}

            \hline
            \multirow{6}{*}{UAV-Human\cite{b2}} 
                & \multirow{1}{*}{\cite{b82}} 
                    & Proposed Novel Method  
                        & \begin{tabular}{c|c|c}
                                Precision & Recall & F1 Score \\ 
                                \hline
                                0.49 & 0.49 & 0.48 \\ 
                            \end{tabular} \\
                \cline{2-4}
    
                & \multirow{3}{*}{\cite{b83}} 
                    & CLIP\cite{b85} 
                        & \begin{tabular}{c}
                                Top1/Top5\tablefootnote{Top1/Top5: The authors of \cite{b83}, employed the metric of Top1/Top5 for measuring accuracy in single-label classification.} (Filtering ratio 90\%) \\ 
                                \hline
                                1.79 / 7.05 \\ 
                            \end{tabular} \\
                    \cline{3-4}       
                    & & ViFi CLIP\cite{b86} 
                        & \begin{tabular}{c}
                                Top1/Top5 (Filtering ratio 90\%) \\ 
                                \hline
                                4.67 / 15.18 \\ 
                            \end{tabular} \\ 
                \cline{2-4}
    
                & \multirow{3}{*}{\cite{b84}} 
                    & 2s-MS\&TA-HGCN-FC (Novel method)
                        & \begin{tabular}{c|c}
                                CSv1 & CSv2 \\ 
                                \hline
                                44.33 & 70.69 \\ 
                            \end{tabular} \\
                    \cline{3-4}       
                    & & 4s-MS\&TA-HGCN-FC (Novel method) 
                        & \begin{tabular}{c|c}
                                CSv1 & CSv2 \\ 
                                \hline
                                45.72 & 71.84 \\ 
                            \end{tabular} \\ 
                    \cline{3-4}
                    & & FR-AGCN\cite{b87} 
                        & \begin{tabular}{c|c}
                                CSv1 & CSv2 \\ 
                                \hline
                                43.98 & 69.5 \\ 
                            \end{tabular} \\
                \cline{1-4}


            \hline
            \multirow{6}{*}{UAVDark135\cite{b13}} 
                & \multirow{3}{*}{\cite{b92}} 
                    & DCPT 
                        & \begin{tabular}{c|c|c}
                                Success Rate & Precision & Normalized Precision \\ 
                                \hline
                                0.577 & 0.703 & 0.701 \\
                            \end{tabular} \\
                    \cline{3-4}
                    & & DIMP50-SCT 
                        & \begin{tabular}{c|c|c}
                                Success Rate & Precision & Normalized Precision \\
                                \hline
                                0.562 & 0.717 & 0.71 \\
                            \end{tabular} \\
                    \cline{3-4}
                    & & DIMP18\cite{b116} 
                        & \begin{tabular}{c|c|c}
                                Success Rate & Precision & Normalized Precision \\
                                \hline
                                0.542 & 0.702 & 0.69 \\
                            \end{tabular} \\
                \cline{2-4}
                & \multirow{3}{*}{\cite{b93}} 
                    & DL+SiamAPN 
                        & \begin{tabular}{c|c}
                                Success Rate & Precision \\ 
                                \hline
                                0.389 & 0.516 \\
                            \end{tabular} \\
                    \cline{3-4}       
                    & & SiamAPN\cite{b100}
                        & \begin{tabular}{c|c}
                                Success Rate & Precision \\ 
                                \hline
                                0.3 & 0.424 \\
                            \end{tabular} \\
                    \cline{3-4}       
                    & & DL+DIMP50
                        & \begin{tabular}{c|c}
                                Success Rate & Precision \\ 
                                \hline
                                0.544 & 0.7 \\
                            \end{tabular} \\
                    \cline{3-4}       
                    & & DIMP50\cite{b115}
                        & \begin{tabular}{c|c}
                                Success Rate & Precision \\ 
                                \hline
                                0.526 & 0.672 \\
                            \end{tabular} \\
                \cline{1-4}

            \hline
            \multirow{6}{*}{VRAI\cite{b5}} 
                & \multirow{3}{*}{\cite{b113}} 
                    & Proposed Novel Method 
                        & \begin{tabular}{c|c}
                                mAP & R-1 Accuracy \\ 
                                \hline
                                0.828 & 0.844 \\ 
                            \end{tabular} \\
                    \cline{3-4}       
                    & & TransReID\cite{b114}
                        & \begin{tabular}{c|c}
                                mAP & R-1 Accuracy \\ 
                                \hline
                                0.814 & 0.826 \\ 
                            \end{tabular} \\  
                \cline{2-4}
    
                & \multirow{3}{*}{\cite{b114}} 
                    & RotTrans
                        & \begin{tabular}{c|c}
                                mAP & R-1 Accuracy \\ 
                                \hline
                                0.848 & 0.838 \\ 
                            \end{tabular} \\
                    \cline{3-4}       
                    & & TransReID 
                        & \begin{tabular}{c|c}
                                mAP & R-1 Accuracy \\ 
                                \hline
                                0.786 & 0.803 \\ 
                            \end{tabular} \\ 
                \cline{1-4}
    \end{tabular}
    }
    \label{tab:performance2}
\end{table}

\begin{table}[htbp]
    \centering
    \caption{Performance Metrics and Results for Different Datasets and Methods}
    \resizebox{\textwidth}{!}{%
    \begin{tabular}{|c|c|c|c|}
        \hline
        \textbf{Dataset Name} & \textbf{Reference} & \textbf{Methods} & \textbf{Performance} \\ 

            \hline
            \multirow{6}{*}{UAV-Gesture\cite{b11}} 
                & \multirow{3}{*}{\cite{b95}} 
                    & Novel Multifeature+CNN method 
                        & \begin{tabular}{c}
                                Accuracy\\ 
                                \hline
                                0.95 \\
                            \end{tabular} \\
                    \cline{3-4}       
                    & & P-CNN\cite{b117}
                        & \begin{tabular}{c}
                                Accuracy \\ 
                                \hline
                                0.91 \\
                            \end{tabular} \\
                    \cline{3-4}       
                    & & MLP\_7j\cite{b94}
                        & \begin{tabular}{c}
                                Accuracy \\ 
                                \hline
                                0.94 \\
                            \end{tabular} \\
                    \cline{2-4}
                & \multirow{3}{*}{\cite{b94}} 
                    & DD-Net\_7j\cite{b118}
                        & \begin{tabular}{c}
                                Accuracy \\ 
                                \hline
                                0.915 \\
                            \end{tabular} \\
                    \cline{3-4}       
                    & & P-CNN
                        & \begin{tabular}{c}
                                Accuracy \\ 
                                \hline
                                0.919 \\
                            \end{tabular} \\
                    \cline{3-4}       
                    & & MLP\_7j
                        & \begin{tabular}{c}
                                Accuracy \\ 
                                \hline
                                0.948 \\
                            \end{tabular} \\
                \cline{1-4}
            \hline
            \multirow{6}{*}{UAVid\cite{b4}} 
                & \multirow{3}{*}{\cite{b98}} 
                    & BANet 
                        & \begin{tabular}{c}
                                mIoU(\%)\\ 
                                \hline
                                64.6 \\
                            \end{tabular} \\
                    \cline{3-4}       
                    & & MSD benchmark\cite{b4}
                        & \begin{tabular}{c}
                                mIoU(\%) \\ 
                                \hline
                                57.0 \\
                            \end{tabular} \\
                    \cline{2-4}
                & \multirow{3}{*}{\cite{b103}} 
                    & A²-FPN
                        & \begin{tabular}{c}
                                mIoU(\%) \\ 
                                \hline
                                65.7 \\
                            \end{tabular} \\
                    \cline{3-4}       
                    & & MSD benchmark
                        & \begin{tabular}{c}
                                mIoU(\%) \\ 
                                \hline
                                57.0 \\
                            \end{tabular} \\
                \cline{2-4}
                & \multirow{3}{*}{\cite{b104}} 
                    & UNetFormer
                        & \begin{tabular}{c}
                                mIoU(\%) \\ 
                                \hline
                                67.8 \\
                            \end{tabular} \\
                    \cline{3-4}       
                    & & ABCNet
                        & \begin{tabular}{c}
                                mIoU(\%) \\ 
                                \hline
                                63.8 \\
                            \end{tabular} \\
                    \cline{3-4}       
                    & & BANet
                        & \begin{tabular}{c}
                                mIoU(\%) \\ 
                                \hline
                                64.6 \\
                            \end{tabular} \\
                    \cline{3-4}       
                    & & BoTNet
                        & \begin{tabular}{c}
                                mIoU(\%) \\ 
                                \hline
                                63.2 \\
                            \end{tabular} \\
                \cline{2-4}
                & \multirow{3}{*}{\cite{b107}} 
                    & MSD benchmark
                        & \begin{tabular}{c|c}
                                mIoU(\%) & FPS \\ 
                                \hline
                                57.0 & 1.00 \\
                            \end{tabular} \\
                    \cline{3-4}       
                    & & BiSeNet\cite{b119}
                        & \begin{tabular}{c|c}
                                mIoU(\%) & FPS \\ 
                                \hline
                                61.5 & 11.08 \\
                            \end{tabular} \\
                    \cline{3-4}       
                    & & CAN
                        & \begin{tabular}{c|c}
                                mIoU(\%) & FPS \\ 
                                \hline
                                63.5 & 15.14 \\
                            \end{tabular} \\
                \cline{1-4}

            \hline
            \multirow{6}{*}{VERI-Wild\cite{b15}} 
                & \multirow{2}{*}{\cite{b108}} 
                    & FDA-Net\cite{b120} 
                        & \begin{tabular}{c|c|c}
                                mAP(small) & mAP(medium) & mAP(large) \\ 
                                \hline
                                0.351 & 0.298 & 0.228 \\ 
                            \end{tabular} \\
                    \cline{3-4}
                    & & PVEN 
                        & \begin{tabular}{c|c|c}
                                mAP(small) & mAP(medium) & mAP(large) \\ 
                                \hline
                                0.825 & 0.77 & 0.697 \\ 
                            \end{tabular} \\
                \cline{2-4}
    
                & \multirow{3}{*}{\cite{b109}} 
                    & MLSL\cite{b121} 
                        & \begin{tabular}{c|c}
                                mAP(large) & R-1 accuracy (large) \\ 
                                \hline
                                0.366 & 0.775 \\ 
                            \end{tabular} \\
                    \cline{3-4}       
                    & & FastReID 
                        & \begin{tabular}{c|c}
                                mAP(large) & R-1 accuracy (large) \\ 
                                \hline
                                0.773 & 0.925 \\ 
                            \end{tabular} \\ 
                \cline{2-4}
    
                & \multirow{3}{*}{\cite{b110}} 
                    & GiT
                        & \begin{tabular}{c|c}
                                mAP(T10000) & R-1 accuracy (T10000) \\
                                \hline
                                 0.675 & 0.854 \\ 
                            \end{tabular} \\
                    \cline{3-4}       
                    & & PCRNet\cite{b122} 
                        & \begin{tabular}{c|c}
                                mAP(T10000) & R-1 accuracy (T10000) \\ 
                                \hline
                                0.671 & 0.85 \\ 
                            \end{tabular} \\ 
                \cline{2-4}
    
                & \multirow{3}{*}{\cite{b111}} 
                    & HPGN
                        & \begin{tabular}{c|c}
                                mAP(T10000) & R-1 accuracy (T10000) \\ 
                                \hline
                                0.65 & 0.8268 \\ 
                            \end{tabular} \\
                    \cline{3-4}       
                    & & Triplet Embedding\cite{b123} 
                        & \begin{tabular}{c|c}
                                mAP(T10000) & R-1 accuracy (T10000) \\ 
                                \hline
                                0.516 & 0.699 \\ 
                            \end{tabular} \\ 
                \cline{2-4}
    
                & \multirow{3}{*}{\cite{b112}} 
                    & Baseline\cite{b124}
                        & \begin{tabular}{c|c}
                                mAP(large) & R-1 accuracy (large) \\ 
                                \hline
                                0.65 & 0.95 \\ 
                            \end{tabular} \\
                    \cline{3-4}       
                    & & SAVER 
                        & \begin{tabular}{c|c}
                                mAP(large) & R-1 accuracy (large) \\ 
                                \hline
                                0.677 & 0.958 \\ 
                            \end{tabular} \\ 
                \cline{1-4}


    \end{tabular}
    }
    \label{tab:performance3}
\end{table}

\section{Results and Discussion of Reviewed Papers}
The datasets discussed in this section represent the application of the papers reviewed in this survey. Our analysis of the datasets revealed that KITE, RescueNet, and Biodrone are relatively new and have not been thoroughly investigated in the literature. While one of the datasets we reviewed, ERA, is not very recent, it still lacks the enough amount of study to fully emphasize its potential. The datasets included in our review were selected based on the number of citations their associated papers have received, emphasizing those with higher citation counts. We delved into several papers that make compelling use of the datasets we evaluated. In our examination, we carefully reviewed the details of the analysis of results and experiments conducted by other researchers. These researchers utilized the datasets we assessed as benchmarks and applied various methods. We have included the best results for the methods applied to the datasets we reviewed in this section and in Table \ref{tab:performance}, \ref{tab:performance2} and \ref{tab:performance3}. 

\subsection{AU-AIR}
In their study, Jiahui et al.\cite{b71} selected AU-AIR as a benchmark dataset to create their proposed real-time object detection model, RSSD-TA-LATM-GID, specifically designed for small-scale object detection. The performance of their model surpassed that of YOLOv4\cite{b102} and YOLOv3\cite{b101}. The researchers employed the MobileNetv-SSDLite ensemble approach, which yielded the lowest mean average precision (mAP) score. 

Walambe et al.\cite{b73} employed baseline models on the AU-AIR dataset as one of their evaluative benchmarks. The objective of the study was to demonstrate the attainability of different techniques and ensemble techniques in the detection of objects with varying scales. The baseline technique yielded the highest performance, with a mean average precision (mAP) score of 6.63\%. This outcome was achieved by employing color-augmentation on the dataset. The performance metrics for the ensemble methods YOLO+RetinaNet and RetinaNet+SSD were found to be 3.69\% and 4.03\%, respectively. The authors Saeed et al.\cite{b72} made modifications to the architecture of the CenterNet model by using other Convolutional Neural Networks (CNNs) as backbones, such as resnet18, hourglass-104, resnet101, and res2net101. Among all the CNNs as backbone. The findings are presented in Table \ref{tab:performance}. 

Gupta and Verma in their paper \cite{b74} utilized the AU-AIR data as a reference point, employing a range of advanced models to achieve precise and automated detection and classification of road traffic. The YOLOv4 model achieved the highest mean average precision (mAP) score of 25.94\% on the AU-AIR dataset. The Faster R-CNN and YOLOv3 models achieved the second and third highest maximum average precision (mAP) scores, with values of 13.77\% and 13.33\% respectively.

\subsection{FOR-instance}
Bountos et. al. extensively utilized the "FOR-Instance" dataset in their study, \cite{b80}, while introducing their innovative approach FoMo-Net. The dataset was utilized to analyze point cloud representations obtained from LiDAR sensors in order to gain a deeper understanding of tree geometry. Existing baseline techniques such as PointNet, PointNet++, and Point Transformer were employed to accomplish these objectives on aerial modality. The corresponding findings are presented in Table \ref{tab:performance2}. In a separate paper, Zhang et. al.\cite{b81} used the "FOR-instance" dataset to train their proposed HFC algorithm and compare its performance with other established approaches. The authors utilized several techniques and ensemble approaches (Xing2023, HFC+Xing2023, HFC+Mean Shift, HFC) on several forest types (CULS, NIBIO, NIBIO2, SCION, RMIT, TUWIEN) shown in the FOR-instance dataset. Among all the methods, HFC demonstrated superior performance. The optimal outcomes achieved by the HFC approach on various forest types represented in the FOR-instance dataset are presented in Table \ref{tab:performance2}.

\subsection{UAV-Assistant}
Albanis et al. used the UAV-Assistant dataset as benchmark for their research, \cite{b88}. They conducted a comparative analysis of BPnP\cite{b89} and HigherHRNet's\cite{b90} 6DOF object pose estimation using several different criteria. Analysis revealed that loss functions play a crucial role in posture estimation. Specifically, the l\_p loss function outperformed the l\_h loss function, particularly in the case of the M2ED drone, resulting in improved accuracy metrics. HigherHRNet demonstrated greater performance compared to HRNet\cite{b91} on smaller objects such as the Tello drone, but not on the M2ED drone, indicating its potential superiority under smaller object classifications. Their analysis of qualitative heatmaps revealed that the l\_p loss function performed better than the Gaussian-distributed l\_h model in accurately locating keypoints. Table \ref{tab:performance2} displays the accuracy metrics (ACC2 and ACC5) obtained from the research conducted by Albanis and his colleagues. In the case of BPnP, we have included the accuracy for both M2ED and Tello drones respectively, as they achieved the highest accuracy outcomes. Regarding HRNet and HigherHRNet, they achieved the best accuracy specifically for M2ED. 

\subsection{AIDER}
The AIDER dataset has been utilized as a benchmark by Alrayes et al. and the authors of AIDER in developing their innovative method, "EmergencyNet." In their paper, \cite{b75} various pre-trained models were applied to the AIDER dataset, with the best F1 accuracy achieved using VGG16 (96.4\%) and ResNet50 (96.1\%). However, the memory consumption for VGG16 and ResNet50 was quite high, at 59.39MB and 96.4MB respectively. However, EmergencyNet achieved 95.7\% F1 accuracy with only 0.368MB of RAM. ResNet50 had nearly 24 million parameters, while VGG16 had 14.8 million. Alrayes et al. benchmarked their AISCC-DE2MS model with AIDER. They found that their algorithm outperformed the genetic, cat-swarm, and artificial bee colony algorithms. MSE and PSNR were utilized to evaluate. These methods were used to compare five photos to evaluate the model's performance. The best result from the five photos is shown in Table \ref{tab:performance}.

\subsection{DarkTrack2021}
Changhong Fu and his team utilized the DarkTrack2021 benchmark as a foundation for developing the Segment Anything Model (SMA) powered framework SAM-DA. Their research \cite{b77} focused on effectively addressing illumination variation and low ambient intensity. They conducted a comparative analysis between their model and various methods, particularly the Baseline tracker UDAT\cite{b79} method. Their novel approach outperformed the Baseline UDAT method, achieving substantial improvements of 7.1\% in illumination variation and 7.8\% in low ambient intensity. The authors evaluated 15 state-of-the-art trackers and found that SAM-DA demonstrated the most promising results. Additionally, Changhong Fu delved into Siamese Object Tracking in their another study \cite{b78}, highlighting the significance of UAVs in visual object tracking. They also leveraged the DarkTrack2021 datasets as a benchmark to assess model performance in low-illumination conditions, with detailed results and the applied models presented in Table \ref{tab:performance}.

\subsection{UAV-Human}
Azmat et al.\cite{b82} address UAV-captured data-based human action recognition (HAR) challenges and approaches in their UAV-Human dataset research. Azmat et al. evaluated their HAR system on 67,428 video sequences of 119 people in various contexts from the UAV-Human dataset. The approach has a mean accuracy of 48.60\% across eight action classes, indicating that backdrops, occlusions, and camera motion hinder human movement recognition in this dataset. Lin et al.\cite{b83} studied text bag filtering techniques for model training, emphasizing data quality. Their ablation study indicated that text bag filtering ratio influences CLIP matching accuracy and zero-shot transfer performance. Filtering training data improved model generalization, especially in unsupervised learning. Huang et al.\cite{b84} evaluated the 4s-MS\&TA-HGCN-FC skeleton-based action recognition model on the UAV-Human dataset. The model achieved 45.72\% accuracy on the CSv1 benchmark and 71.84\% on the CSv2 test, surpassing previous state-of-the-art techniques. They found that their technique can manage UAV-captured data's viewpoints, motion blurring, and resolution changes.

\subsection{UAVDark135}
Zhu et al.\cite{b92} and Ye et al.\cite{b93} used the UAVDark135 dataset to evaluate their strategies for increasing low-light tracking performance. The Darkness Clue-Prompted Tracking (DCPT) approach by Zhu et al. showed considerable gains, reaching a 57.51\% success rate on UAVDark135. A 1.95\% improvement over the base tracker demonstrates the effectiveness of including darkness clues. Additionally, DCPT's gated feature aggregation approach increased success score by 2.67\%, making it a reliable nighttime UAV tracking system. Ye et al.'s DarkLighter(DL) approach improved tracking performance on the UAVDark135 dataset. DL improved SimpAPN\cite{b99}\cite{b100} tracker AUC by over 29\% and precision by 21\%. It also worked well across tracking backbones, enhancing precision and success rates in light variation, quick motion, and low resolution circumstances. DL surpassed modern low-light enhancers like LIME by 1.68\% in success rate and 1.45\% in precision.

\subsection{VRAI}
VRAI was utilized to establish a vehicle re-identification baseline. Syeda Nyma Ferdous, Xin Li, and Siwei Lyu \cite{b113} tested their uncertainty-aware multitask learning framework on this dataset and achieved 84.47\% Rank-1 accuracy and 82.86\% mAP. This model's capacity to handle aerial image size and position fluctuations was greatly improved by multiscale feature representation and a Pyramid Vision Transformer (PVT) architecture. 
Shuoyi Chen, Mang Ye, and Bo Du\cite{b114} focused on vehicle ReID using VRAI. RotTrans, a rotation-invariant vision transformer, surpassed current innovative approaches by 3.5\% in Rank-1 accuracy and 6.2\% in mean average precision (mAP). This approach solved UAV-based vehicle ReID challenges that typical pedestrian ReID methods struggle with. The process was further complicated by the need to present results in a certain format for performance evaluation. 

\subsection{UAV-Gesture}
Usman Azmat et al.\cite{b95} and Papaioannidis et al.\cite{b94} utilized the UAV-Gesture dataset to evaluate their recommendations for human action and gesture recognition. They used the UAV-Gesture collection of 119 high-definition RGB movies representing 13 unique motions used to control UAVs. The dataset is ideal for testing recognition systems due to its diversity of views and movement similarities. The Usman Azmat et al. method achieved 0.95 action recognition accuracy on the UAV-Gesture dataset. Mean precision, recall, and F1-score for the system were 0.96, 0.95, and 0.94. Several investigations supported by confusion matrices showed the system's ability to distinguish gestures. Papaioannidis et al. found that their gesture recognition method outperformed DD-Net\cite{b97} and P-CNN\cite{b96} by 3.5\% in accuracy. The authors stressed the need of using 2D skeletal data from movies to boost recognition accuracy. Real-time performance makes their method suitable for embedded AI hardware in dynamic UAV situations.

\subsection{UAVid}
The UAVid dataset has been extensively utilized as a benchmark by several researchers in the development of innovative methods for semantic segmentation in urban environments. Wang et al.\cite{b98} introduced the Bilateral Awareness Network (BANet) and applied it to the UAVid dataset, achieving a notable mean Intersection-over-Union (mIoU) score of 64.6\%. BANet's ability to accurately segment various classes within high-resolution urban scenes was demonstrated through both quantitative metrics and qualitative analysis, outperforming other state-of-the-art models like the MSD benchmark. 

Similarly, Rui Li et al.\cite{b103} proposed the Attention Aggregation Feature Pyramid Network (A²-FPN) and reported significant improvements on the UAVid dataset. A²-FPN achieved the highest mIoU across five out of eight classes, surpassing BANet by 1\% in overall performance. The model's effectiveness was particularly evident in its ability to correctly identify moving vehicles, a challenging task for many segmentation models. 

Libo Wang et al.\cite{b104} introduced the UNetFormer, which further pushed the boundaries of semantic segmentation on the UAVid dataset. Achieving an impressive mIoU of 67.8\%, the UNetFormer outperformed several advanced networks, including ABCNet\cite{b105} and hybrid Transformer-based models like BANet and BoTNet\cite{b106}. The UNetFormer demonstrated a strong ability to handle complex segmentation tasks, particularly in accurately identifying small objects like humans. 

Lastly, Michael Ying Yang et al.\cite{b107} applied the Context Aggregation Network(CAN) to the UAVid dataset, achieving a mIoU score of 63.5\% while maintaining a high processing speed of 15 frames per second (FPS). This model was noted for its ability to maintain consistency in both local and global scene semantics, making it a competitive choice for real-time applications in urban environments.

\subsection{VERI-Wild}
The VERI-Wild dataset has been extensively utilized as a benchmark by several researchers in the development of innovative methods for vehicle re-identification (ReID) in real-world scenarios. Meng et al.\cite{b108} introduced the Parsing-based View-aware Embedding Network (PVEN) and applied it to the VERI-Wild dataset, achieving significant improvements in mean Average Precision (mAP) across small, medium, and large test datasets, with increases of 47.4\%, 47.2\%, and 46.9\%, respectively. PVEN’s ability to perform view-aware feature alignment allowed it to consistently outperform state-of-the-art models, particularly in Cumulative Match Characteristic (CMC) metrics, where it showed a 32.7\% improvement over FDA-Net at rank 1.

Similarly, Lingxiao He et al.\cite{b109} evaluated the FastReID toolbox on the VERI-Wild dataset, highlighting its effectiveness in accurately identifying vehicles across various conditions. FastReID achieved state-of-the-art performance, particularly in Rank-1 accuracy(R1-Accuracy) and mAP, showcasing its robustness in handling the complexities of vehicle ReID tasks in surveillance and traffic monitoring environments.

Fei Shen et al.\cite{b110} applied the GiT method on the VeRi-Wild dataset, securing top performance across all testing subsets, including Test3000(T3000), Test5000(T5000), and Test1000(T1000). The GiT method outperformed the second-place method, PCRNet, by 0.41\% in Rank-1 identification rate and 0.45\% in mAP on the Test1000 subset. The study emphasized the importance of leveraging both global and local features, as GiT demonstrated superior generalization across different datasets and conditions. In a separate study, Fei Shen et al.\cite{b111} developed the Hybrid Pyramidal Graph Network (HPGN) approach, which achieved the highest Rank-1 identification rate among the evaluated methods on the VERI-Wild dataset, so making more contributions to the advancing field of vehicle ReID. The findings highlighted the resilience of HPGN, especially in difficult circumstances such as fluctuating day and night situations, where alternative approaches exhibited a decrease in effectiveness.

Lastly, Khorramshahi et al.\cite{b112} presented a residual generation model that improved mAP by 2.0\% and CMC\@1 by 1.0\% compared to baseline models. The model's reliance on residual information, as indicated by a high alpha value ($\alpha$ = 0.94) which proved crucial in extracting robust features from the dataset. This self-supervised method further proved its adaptability and usefulness in vehicle ReID tasks by showcasing its efficacy on the VERI-Wild dataset.

\section{Conclusion}

In this survey paper, we looked at the current state of UAV datasets, highlighting their various applications, inherent challenges, and future directions. UAV datasets are essential in areas such as disaster management, surveillance, agriculture, environmental monitoring, and human behavior analysis. Advanced machine learning techniques have improved UAV capabilities, enabling more precise data collection and analysis.
Despite their potential, UAV datasets face several challenges, including data quality, consistency, and the need for standardized annotation protocols. Ethical and privacy concerns necessitate strong frameworks to ensure responsible use.
Future research should increase dataset diversity, integrate multimodal data, and improve real-time data processing. Improving annotation quality and promoting collaborative research and open data initiatives will increase dataset utility.
To summarize, UAV datasets are at a critical stage of development, with significant opportunities for technological advancements. Addressing current challenges and focusing on future research directions will result in new discoveries, keeping UAV technology innovative and practical.

\newpage
\bibliographystyle{unsrt}  


\newpage
\appendix
\appendix

\section{Appendix}

The following images were captured from the papers in which they were presented as a new dataset or from the dataset repositories referenced in their paper where they were made available as public dataset repositories. 
\newline
\newline
\newline
\subsection{AIDER}
\hspace{0.2em}
\begin{figure}[h!]
    \centering
    \includegraphics[width=0.98\linewidth, height = 0.6\linewidth]{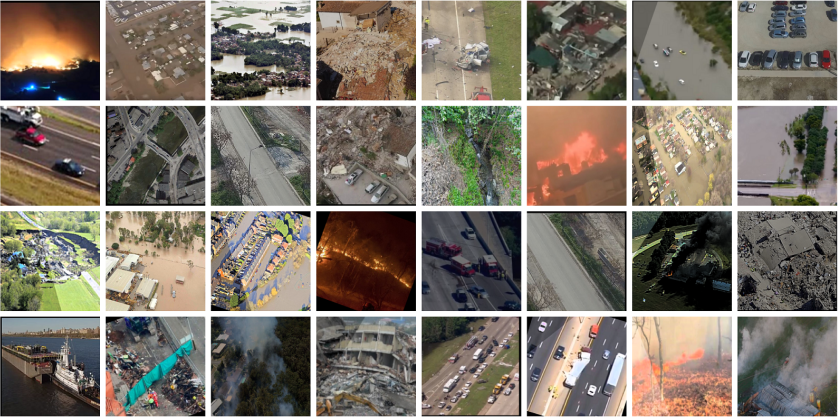}
    \caption{Aerial Image Dataset for Applications in Emergency Response (AIDER): A selection of pictures from the Augmented Database}
    \label{fig:enter-label}
\end{figure}

\newpage
\subsection{BioDrone}
\hspace{0.3em}
\begin{figure}[h!]
    \centering
    \includegraphics[width=1\linewidth]{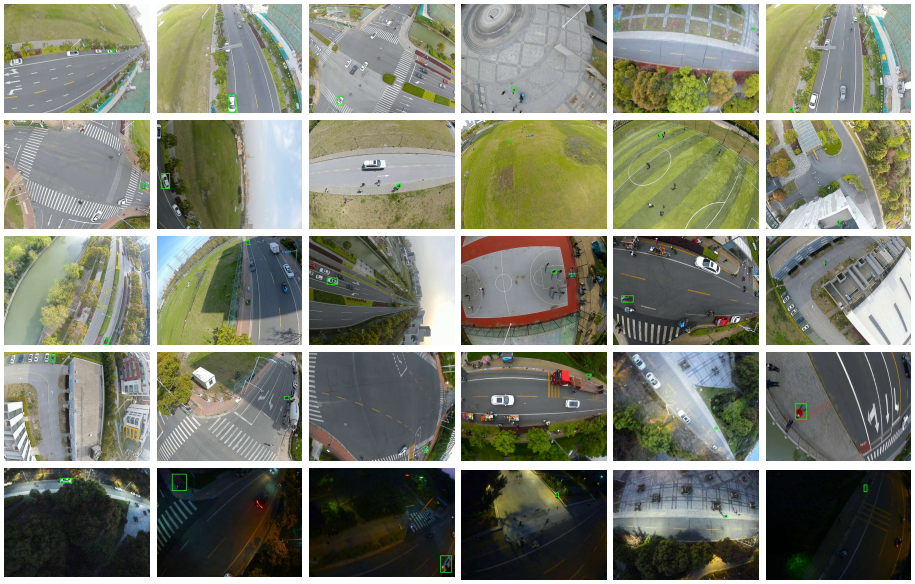}
    \caption{Illustrations of the flapping-wing UAV used for data collection and the representative data of
BioDrone. Different flight attitudes for various scenes under three lighting conditions are included in
the data acquisition process, ensuring that BioDrone can fully reflect the robust visual challenges of the
flapping-wing UAVs.}
    \label{fig:enter-label}
\end{figure}

\newpage
\subsection{ERA}
\hspace{0.2em}
\begin{figure}[h!]
    \centering
    \includegraphics[width=1\linewidth]{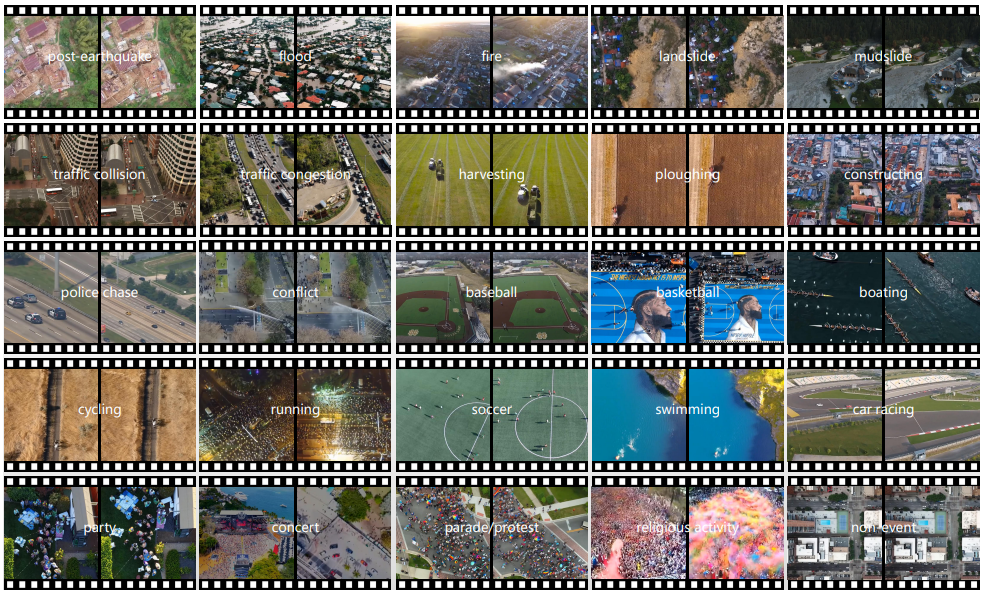}
    \caption{Overview of the ERA dataset. Overall, they have collected 2,864 labeled video snippets for 24 event classes and 1 normal class: post-earthquake,
flood, fire, landslide, mudslide, traffic collision, traffic congestion, harvesting, ploughing, constructing, police chase, conflict, baseball, basketball, boating,
cycling, running, soccer, swimming, car racing, party, concert, parade/protest, religious activity, and non-event. For each class, we show the first (left) and
last (right) frames of a video. Best viewed zoomed in color.}
    \label{fig:enter-label}
\end{figure}

\newpage
\subsection{FOR-instance}
\hspace{0.3em}
\begin{figure}[h!]
    \centering
    \hspace{0.3em}
    \subfigure[Image 1]{\includegraphics[width=0.3\textwidth, height=0.5\textwidth]{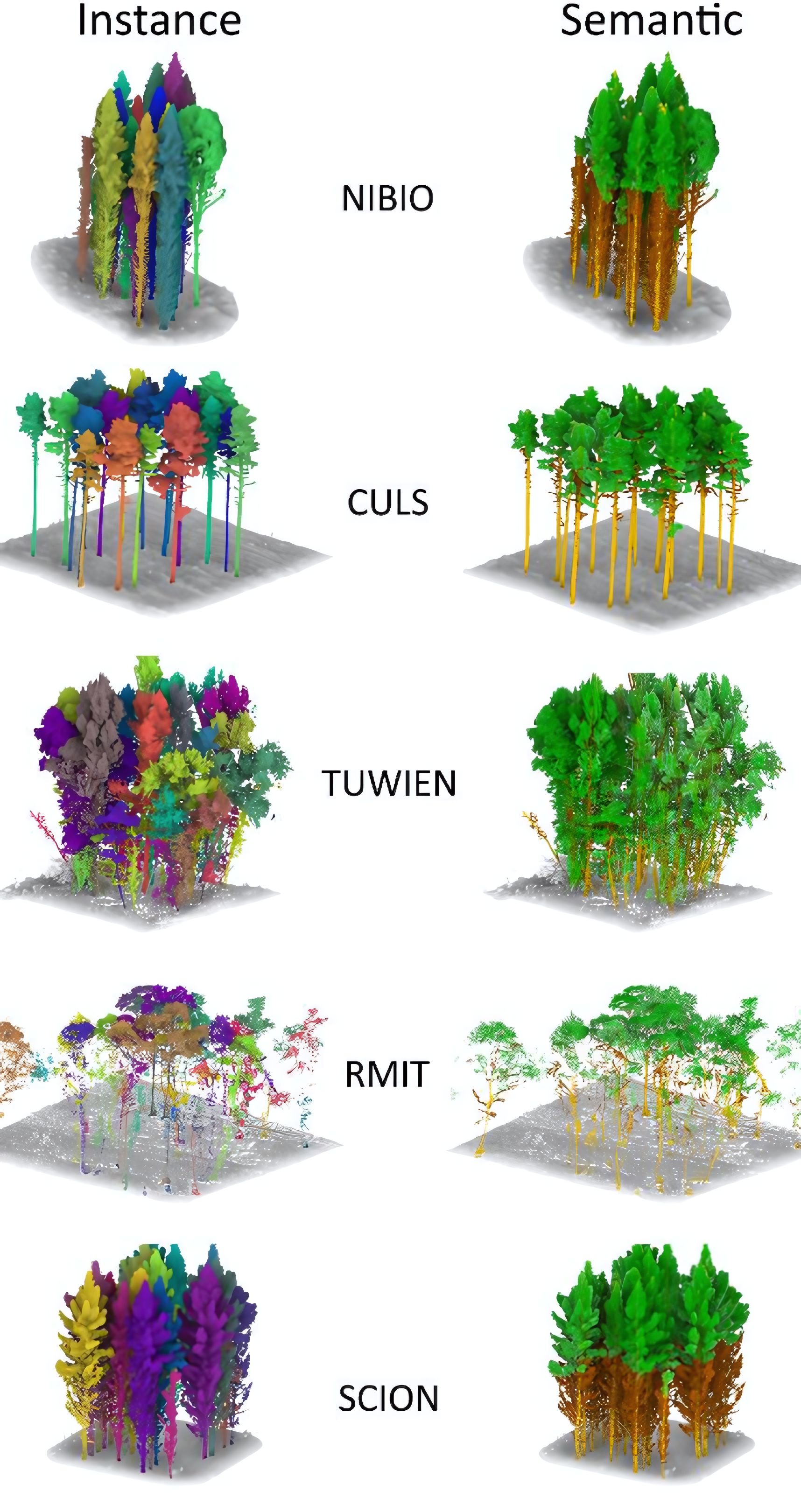}}\hspace{1em}
    \subfigure[Image 2]{\includegraphics[width=0.3\textwidth, height=0.35\textwidth]{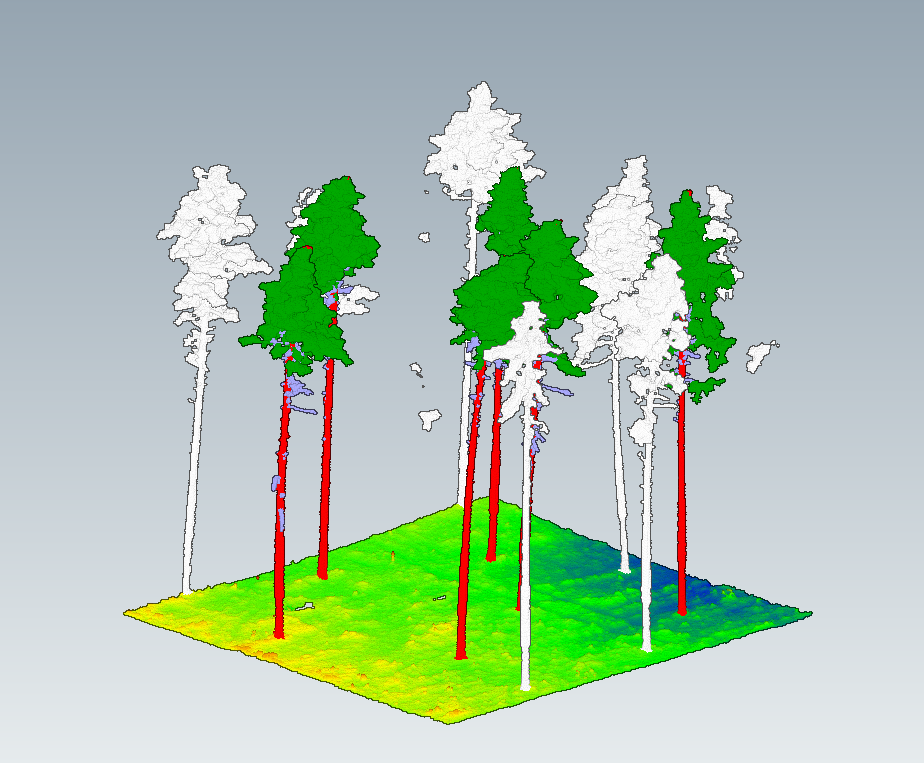}}\hspace{1em}
    \subfigure[Image 3]{\includegraphics[width=0.3\textwidth]{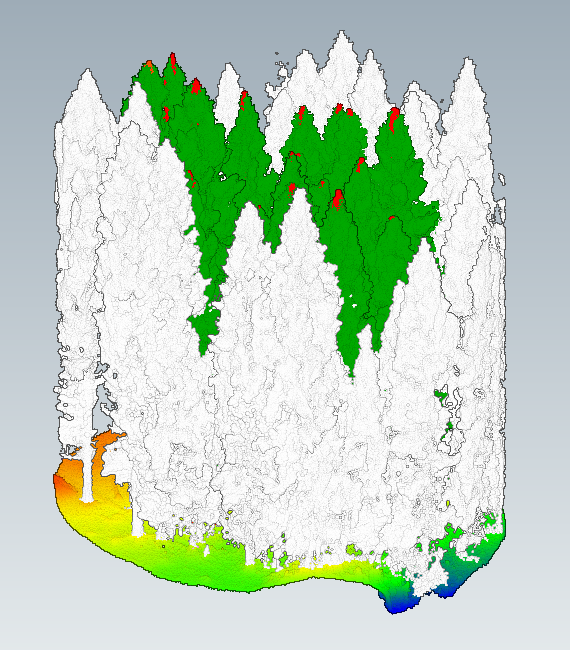}}\\
    \subfigure[Image 4]{\includegraphics[width=0.3\textwidth]{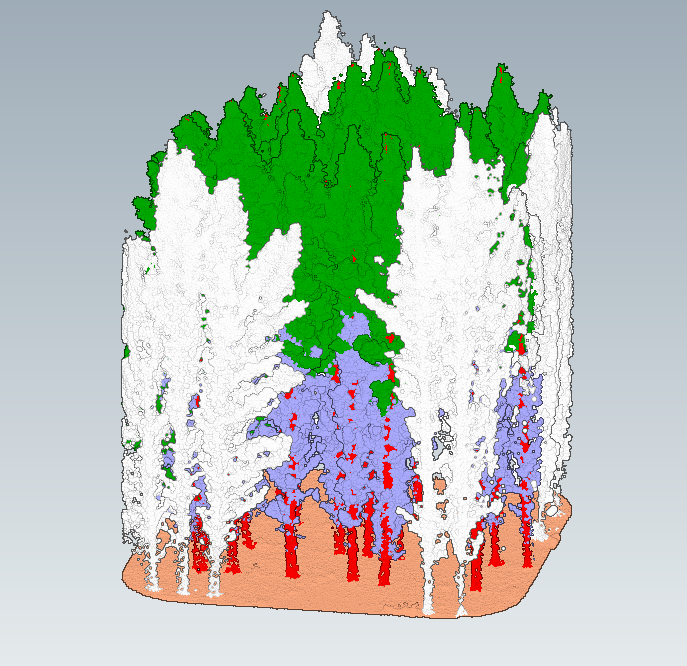}}\hspace{0.5em}
    \subfigure[Image 5]{\includegraphics[width=0.4\textwidth, height=0.25\textwidth]{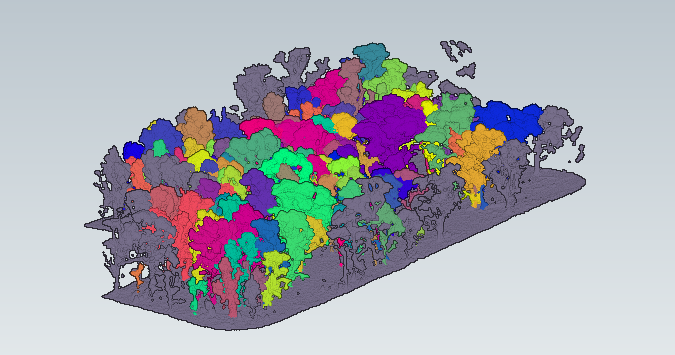}}
    \caption{Samples of the various FOR-instance data collections' instance and semantic annotations.}
    \label{fig:combined_images}
\end{figure}

\newpage
\subsection{UAVDark135}
\hspace{0.3em}
\begin{figure}[h!]
    \centering
    \includegraphics[width=1\linewidth, height = 0.6\linewidth]{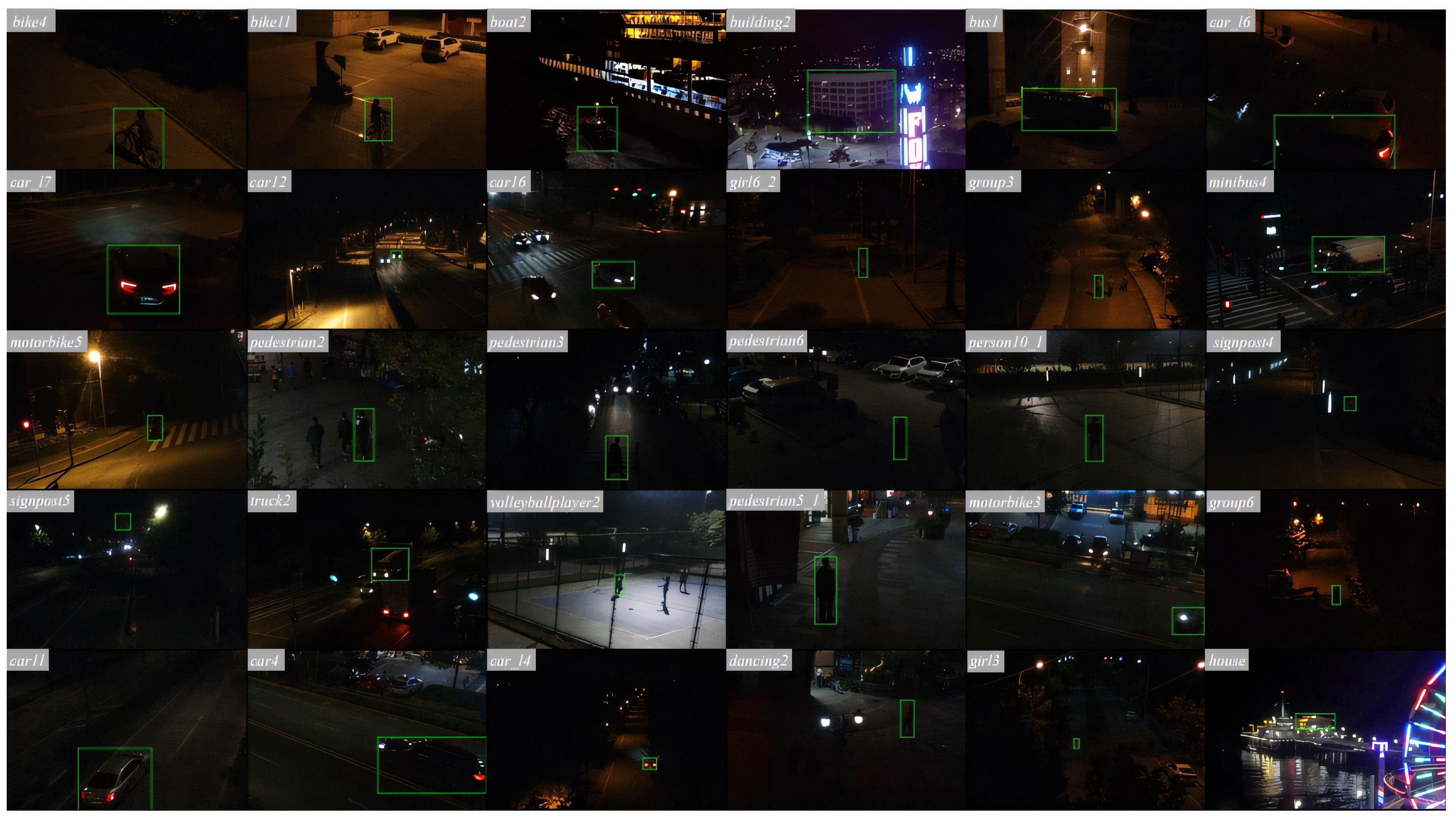}
    \caption{The first frames of representative scenes in newly constructed UAVDark135. Here, target ground-truths are marked out by green boxes and sequence
names are located at the top left corner of the images. Dark special challenges like objects’ unreliable color feature and objects’ merging into the dark can
be seen clearly.}
    \label{fig:enter-label}
\end{figure}

\newpage
\subsection{UAV-Human}
\hspace{0.3em}
\begin{figure}[h!]
    \centering
    \includegraphics[width=0.98\linewidth]{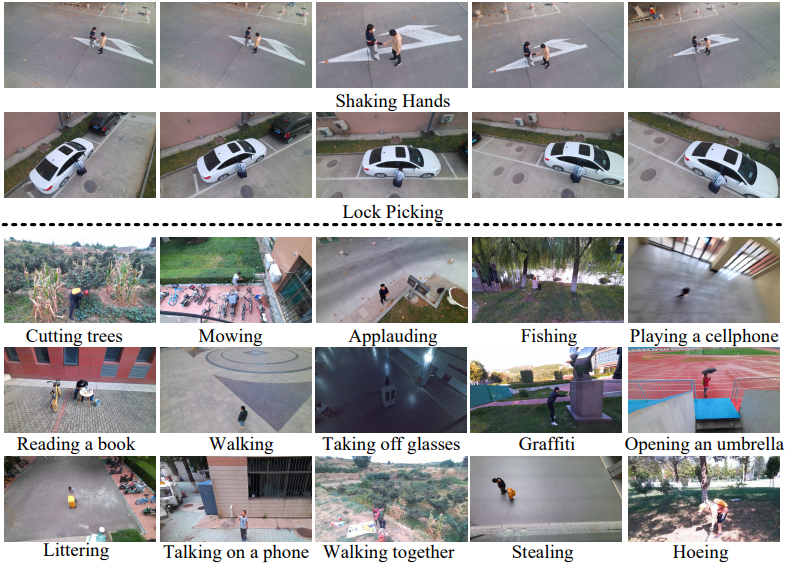}
    \caption{Examples of action videos in UAV-Human dataset. The first and second rows show two video sequences of significant camera motions and view variations, caused by continuously varying flight attitudes, speeds and heights. The last three rows display action samples of the dataset, showing the diversities, e.g., distinct views, various capture sites, weathers, scales, and motion blur.}
    \label{fig:enter-label}
\end{figure}

\newpage
\subsection{UAVid}
\hspace{0.3em}
\begin{figure}[h!]
    \centering
    \includegraphics[width=1\linewidth, height = 0.7\linewidth]{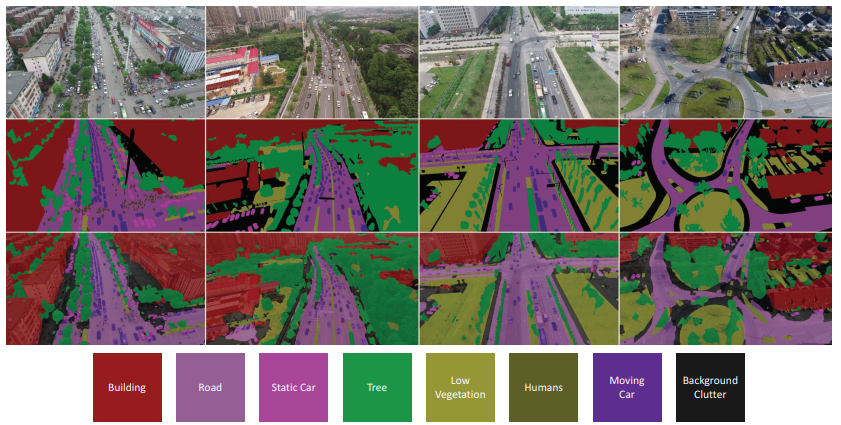}
    \caption{Example images and labels from UAVid dataset. First row shows the images captured by UAV. Second row shows the corresponding ground truth labels. Third row shows the prediction results of MS-Dilation net+PRT+FSO model. The last row shows the labels.}
    \label{fig:enter-label}
\end{figure}

\newpage
\subsection{DarkTrack2021}
\hspace{0.35em}
\begin{figure}[h!]
    \centering
    \includegraphics[width=1\linewidth, height=0.6\linewidth]{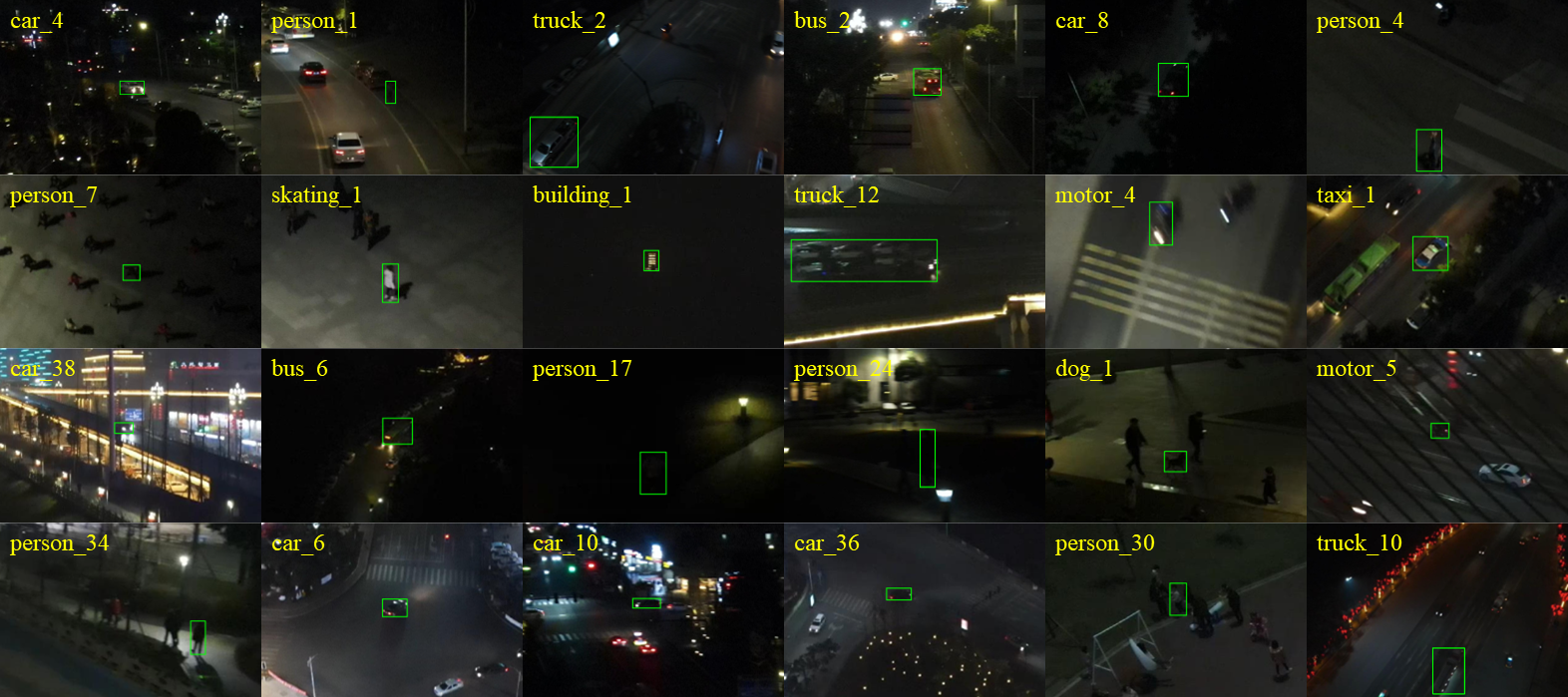}
    \caption{Initial frames of specific sequences from the DarkTrack2021 archive. Objects being tracked are indicated by green boxes, and sequence names are shown in the top left corner of the photos. }
    \label{fig:enter-label}
\end{figure}

\newpage
\subsection{VRAI}
\hspace{0.3em}
\begin{figure}[h!]
    \centering
    \includegraphics[width=0.68\linewidth, height = 0.78\linewidth]{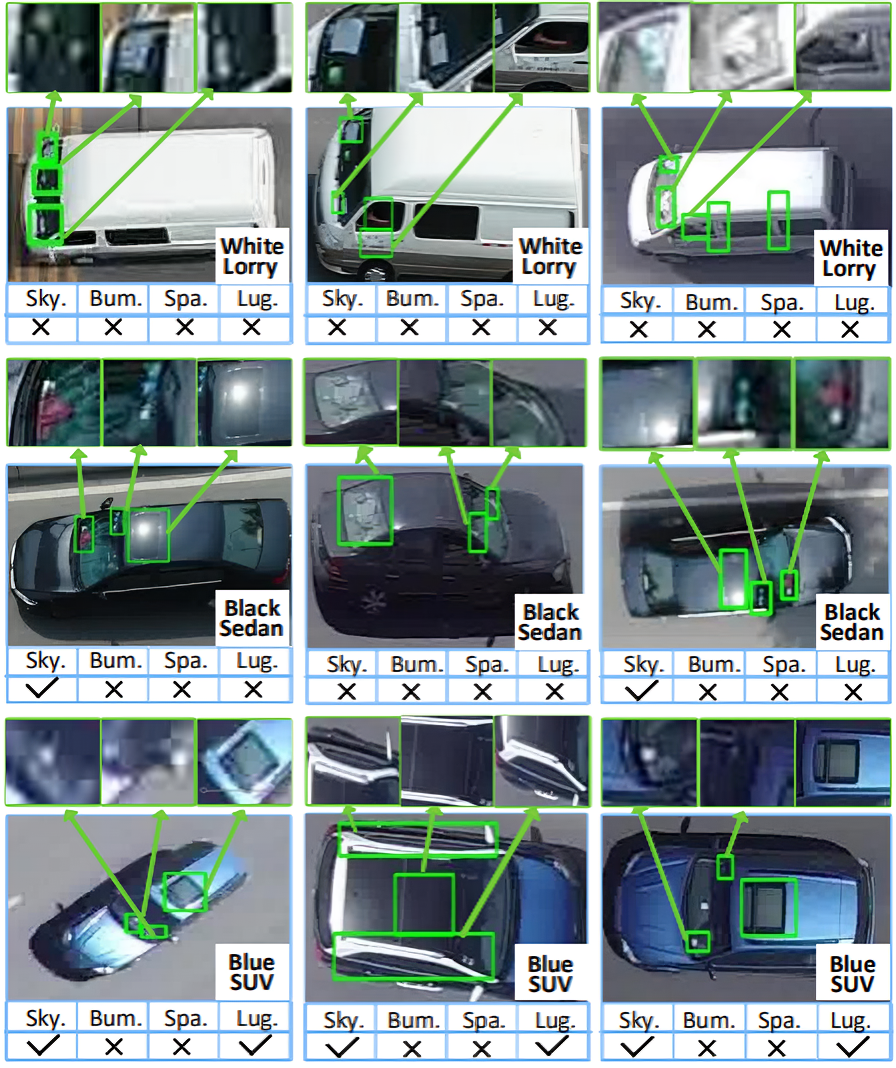}
    \caption{Overview of our gathered dataset for Unmanned Aerial Vehicle (UAV)-based vehicle ReID. In order to facilitate thorough investigation, the authors have included a comprehensive range of information in the dataset, such as color, vehicle type, Skylight (Sky.), Bumper (Bum.), Spare tire (Spa.), Luggage rack (Lug.), and distinguishing components.}
    \label{fig:enter-label}
\end{figure}

\newpage
\subsection{VERI-Wild}
\hspace{0.3em}
\begin{figure}[h!]
    \centering
    \includegraphics[width=1\linewidth]{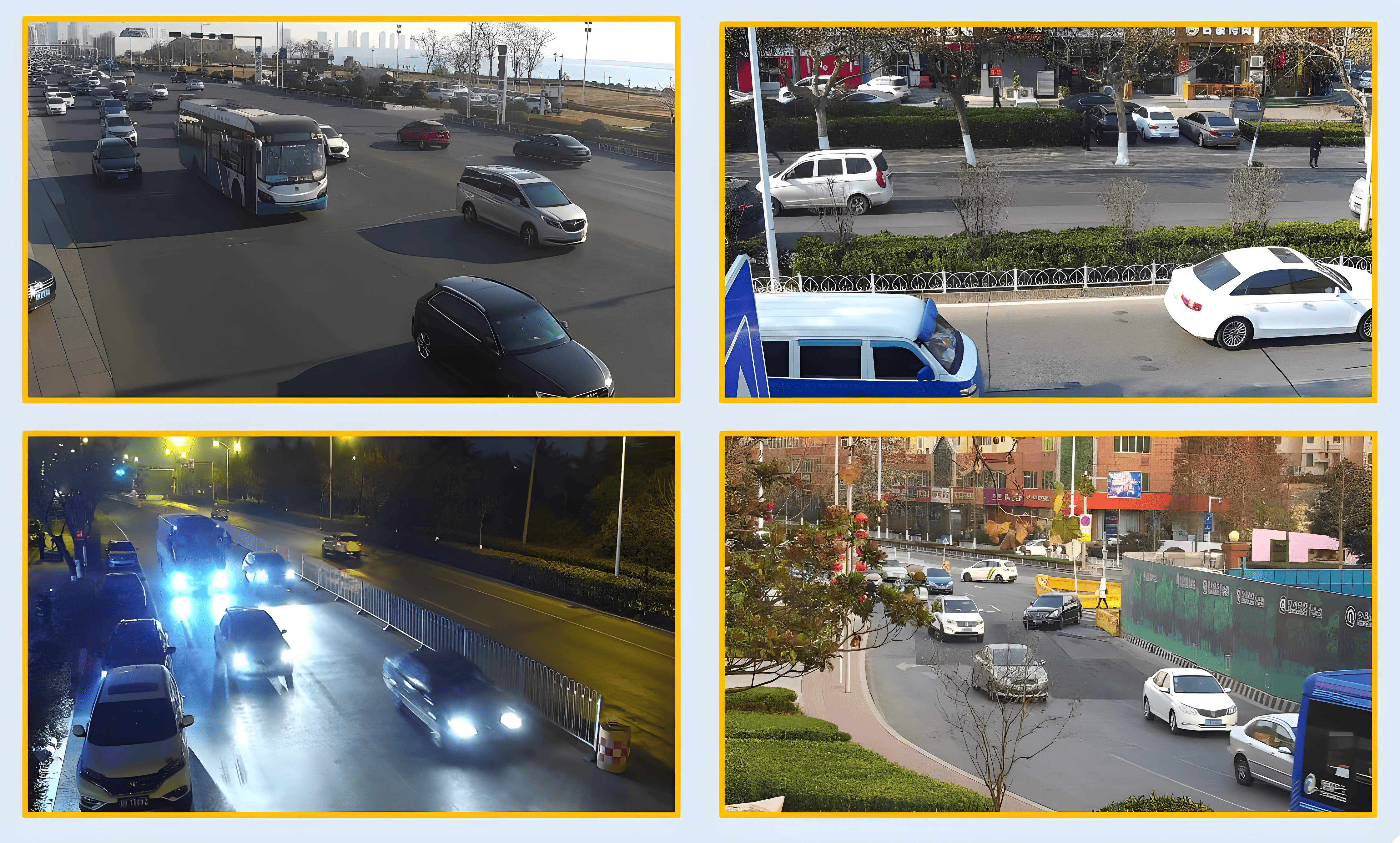}
    \caption{Exemplary photos extracted from the dataset. The dataset is obtained from a comprehensive real video surveillance system including 174 cameras strategically placed around an urban area spanning over 200 square kilometers.}
    \label{fig:enter-label}
\end{figure}

\newpage
\subsection{RescueNet}
\hspace{0.3em}
\begin{figure}[h!]
    \centering
    \includegraphics[width=0.98\linewidth]{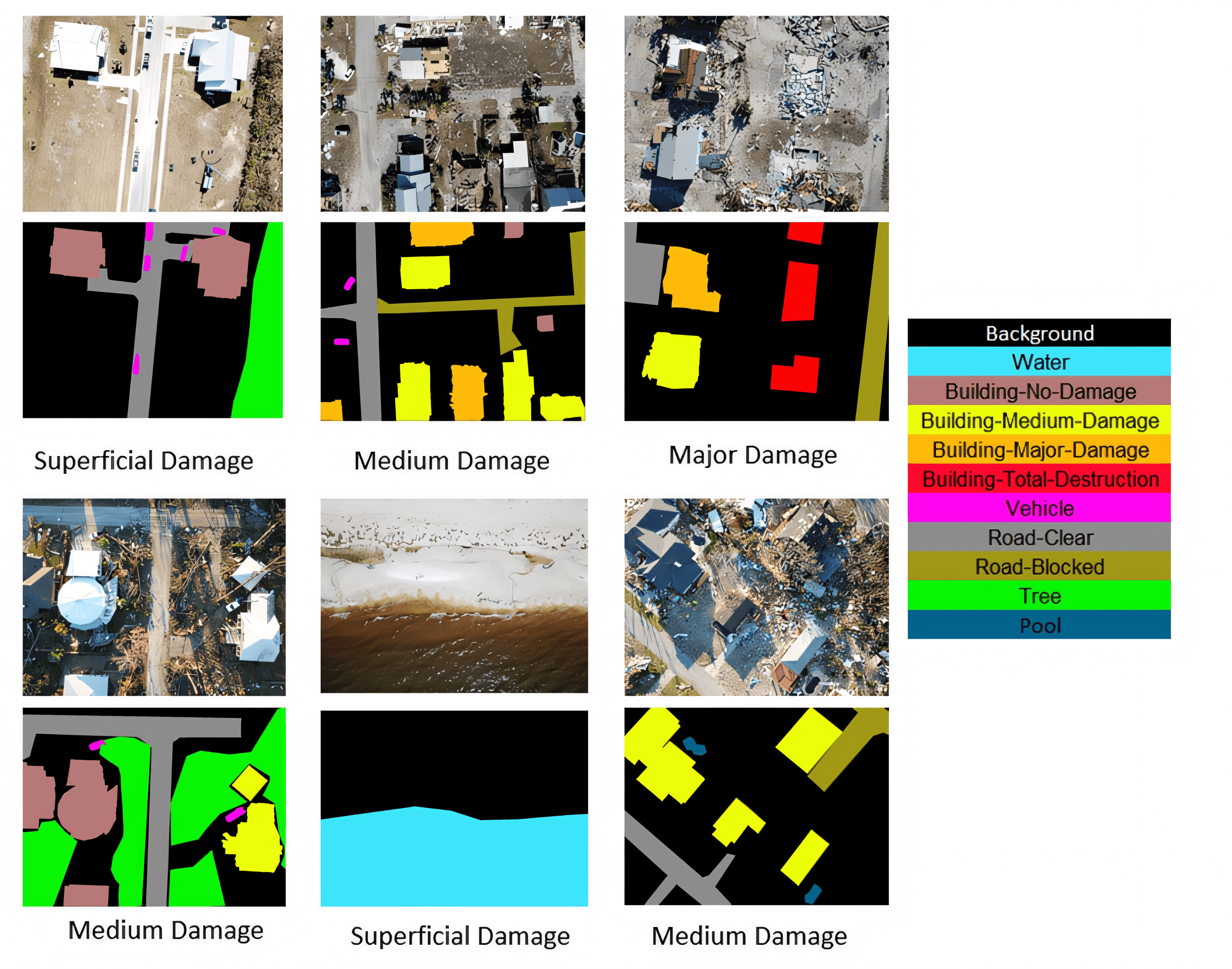}
    \caption{Graphical representation of complex scenes from the RescueNet dataset. The first and third rows display the original photos, while the lower rows provide the associated annotations for both semantic segmentation and image classification functions. Displayed on the right are the 10 classes, each represented by their segmentation color.}
    \label{fig:enter-label}
\end{figure}

\newpage
\subsection{UAV-Assistant}
\hspace{0.3em}
\begin{figure}[h!]
    \centering
    \includegraphics[width=0.73\linewidth]{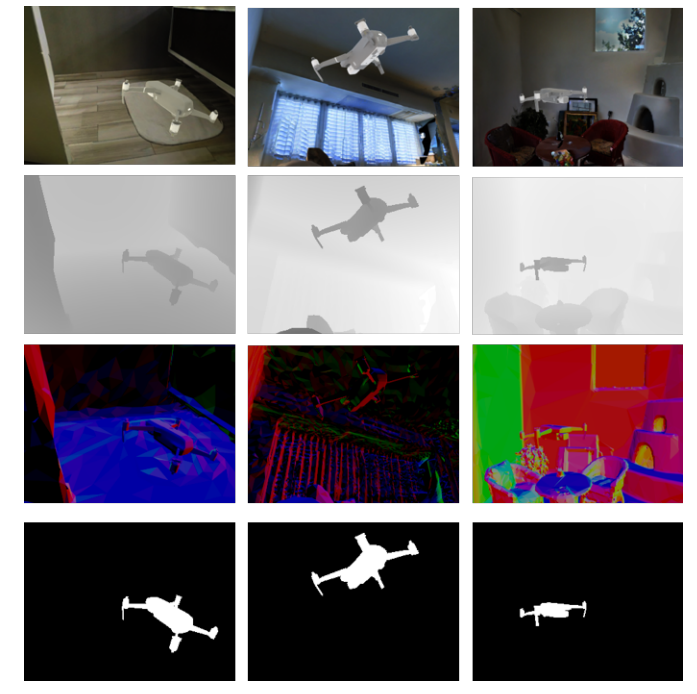}
    \caption{This diagram illustrates the many modalities present in the UAV-Assistant dataset, which consists of a randomly chosen collection of images. The uppermost row displays color photos, the second row displays depth, the third row displays the normal map, and the last row displays flight silhouettes of the drone.}
    \label{fig:enter-label}
\end{figure}

\newpage
\subsection{AU-AIR}
\hspace{0.3em}
\begin{figure}[h!]
    \centering
    \includegraphics[width=0.88\linewidth]{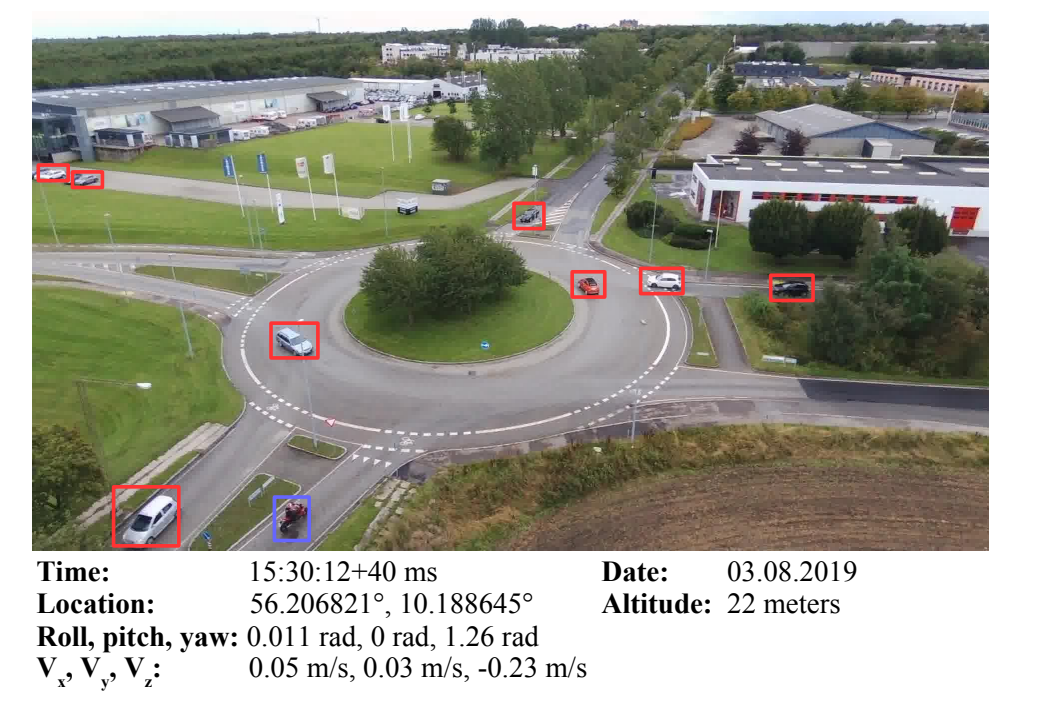}
    \caption{The AU-AIR dataset includes extracted frames that are annotated with object information, time stamp, current location, altitude, velocity of the UAV, and rotation data observed from the IMU sensor. This figure presents an exemplar of it.}
    \label{fig:enter-label}
\end{figure}

\newpage
\subsection{UAV-Gesture}
\hspace{0.3em}
\begin{figure}[h!]
    \centering
    \includegraphics[width=0.8\linewidth]{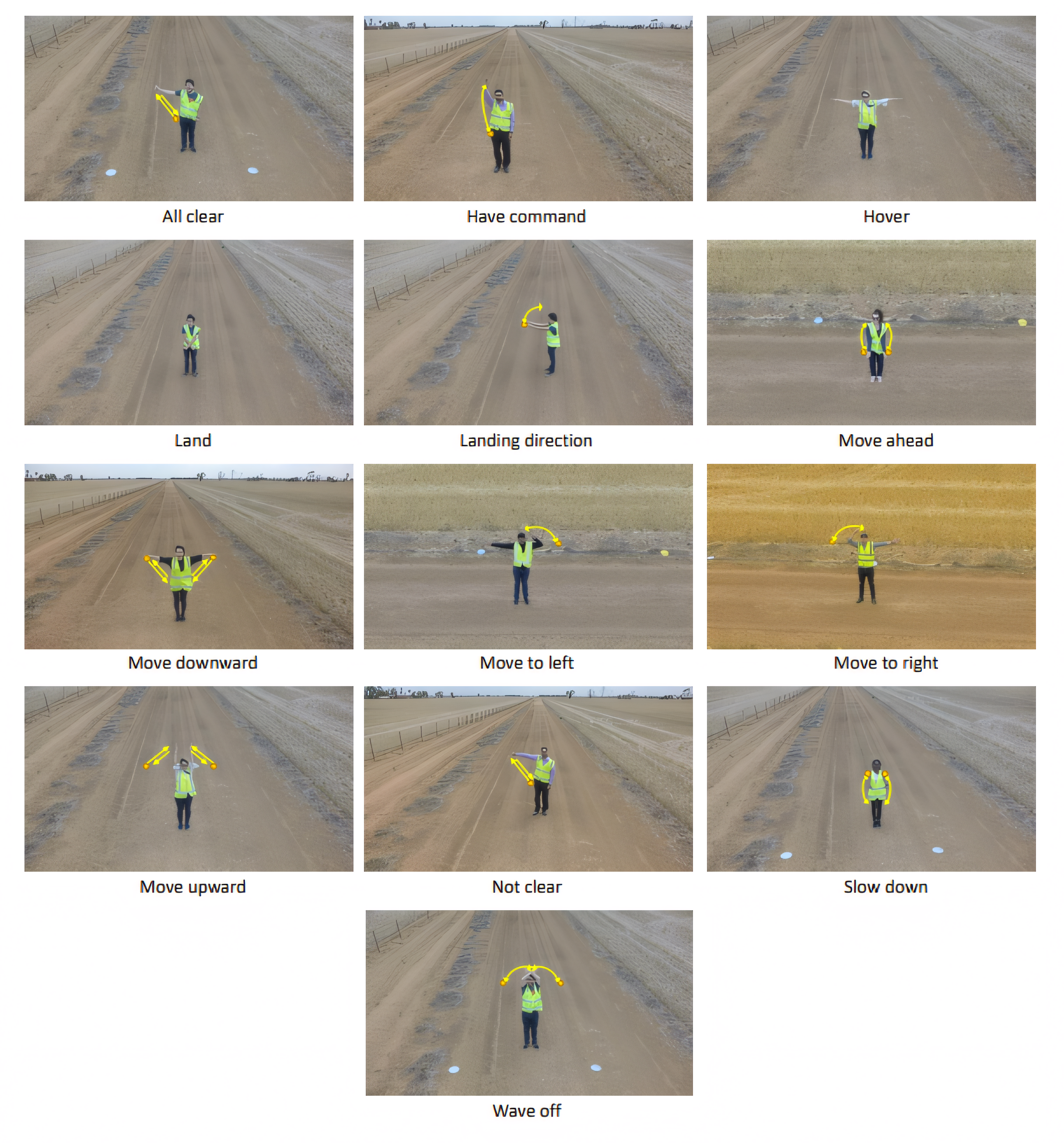}
    \caption{This diagram displays thirteen explicitly chosen gestures, each accompanied by a single picked image. Directions of hand movement are shown by the arrows. The amber color marks serve as approximate indicators of the initial and final locations of the palm for ONE iteration. Neither the Hover nor Land gestures are dynamic gestures.}
    \label{fig:enter-label}
\end{figure}

\newpage
\subsection{Kite}
\hspace{0.3em}
\begin{figure}[h!]
    \centering
    \includegraphics[width=0.86\linewidth]{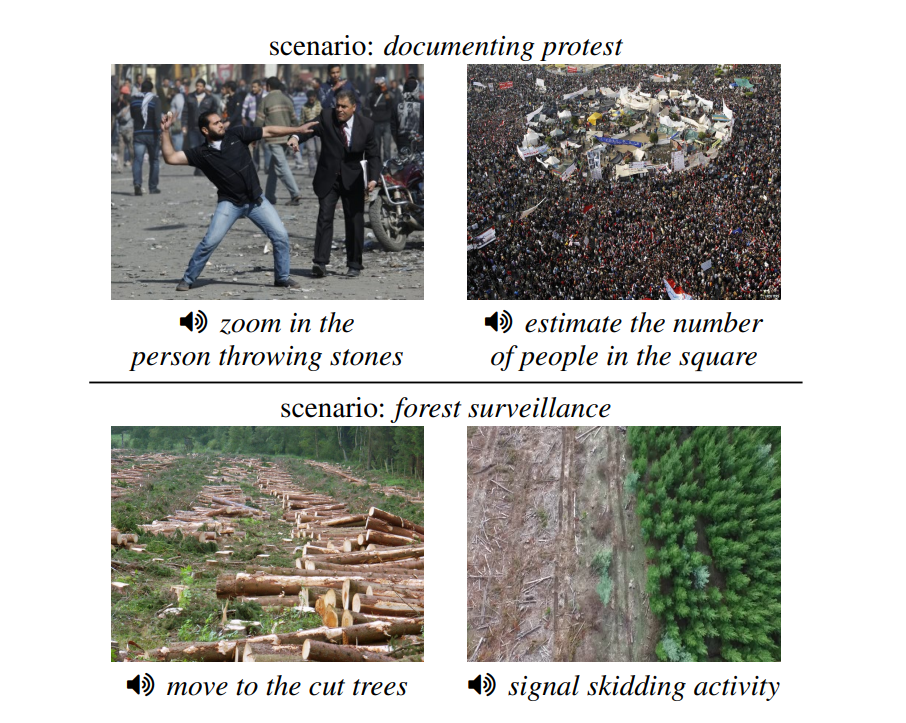}
    \caption{Exemplary commands and visual representations derived from the KITE dataset.}
    \label{fig:enter-label}
\end{figure}

\end{document}